\documentclass{article}

\usepackage[preprint]{neurips_2026}

\usepackage[dvipsnames]{xcolor}
\definecolor{linkColor}{RGB}{38,76,115}  
\definecolor{citecolor}{RGB}{0,105,92}   
\definecolor{urlColor}{RGB}{74,85,104}   
\usepackage[colorlinks=true,linkcolor=linkColor,citecolor=citecolor,filecolor=linkColor,urlcolor=linkColor]{hyperref}

\usepackage[utf8]{inputenc} 
\usepackage[T1]{fontenc}    
\usepackage{hyperref}       
\usepackage{url}            
\usepackage{booktabs}       
\usepackage{amsfonts}       
\usepackage{nicefrac}       
\usepackage{microtype}      
\usepackage{xcolor}         

\usepackage{graphicx}
\usepackage{tabularx}
\usepackage{multirow}
\usepackage{booktabs}
\usepackage{bbding}
\usepackage{caption}
\usepackage{pifont}
\usepackage{stfloats}
\usepackage{bm}
\usepackage{wrapfig}
\usepackage{colortbl}
\usepackage{enumitem}
\usepackage{indentfirst}
\usepackage{algorithm}
\usepackage{algpseudocode}
\usepackage{amsmath}
\usepackage{wrapfig}
\usepackage{booktabs}

\usepackage{fancyhdr} 
\usepackage[most]{tcolorbox} 

\usepackage{caption}
\usepackage{subfigure}

\usepackage{amssymb}

\usepackage{xspace}

\usepackage{duckuments}

\usepackage{makecell}
\usepackage{fontawesome5}

\newcommand{\MLfull}{\faIcon{circle}}
\newcommand{\MLpartial}{\faIcon{adjust}}
\newcommand{\MLempty}{\faIcon[regular]{circle}}

\newcommand{\stylehead}[2]{%
    \makecell[c]{%
        {\large\faIcon{#1}}\\[4pt]
        #2%
    }%
}

\usepackage{algorithm}
\usepackage{algpseudocode}

\title{%
  \raisebox{-0.15em}{%
    \includegraphics[height=1.3em]{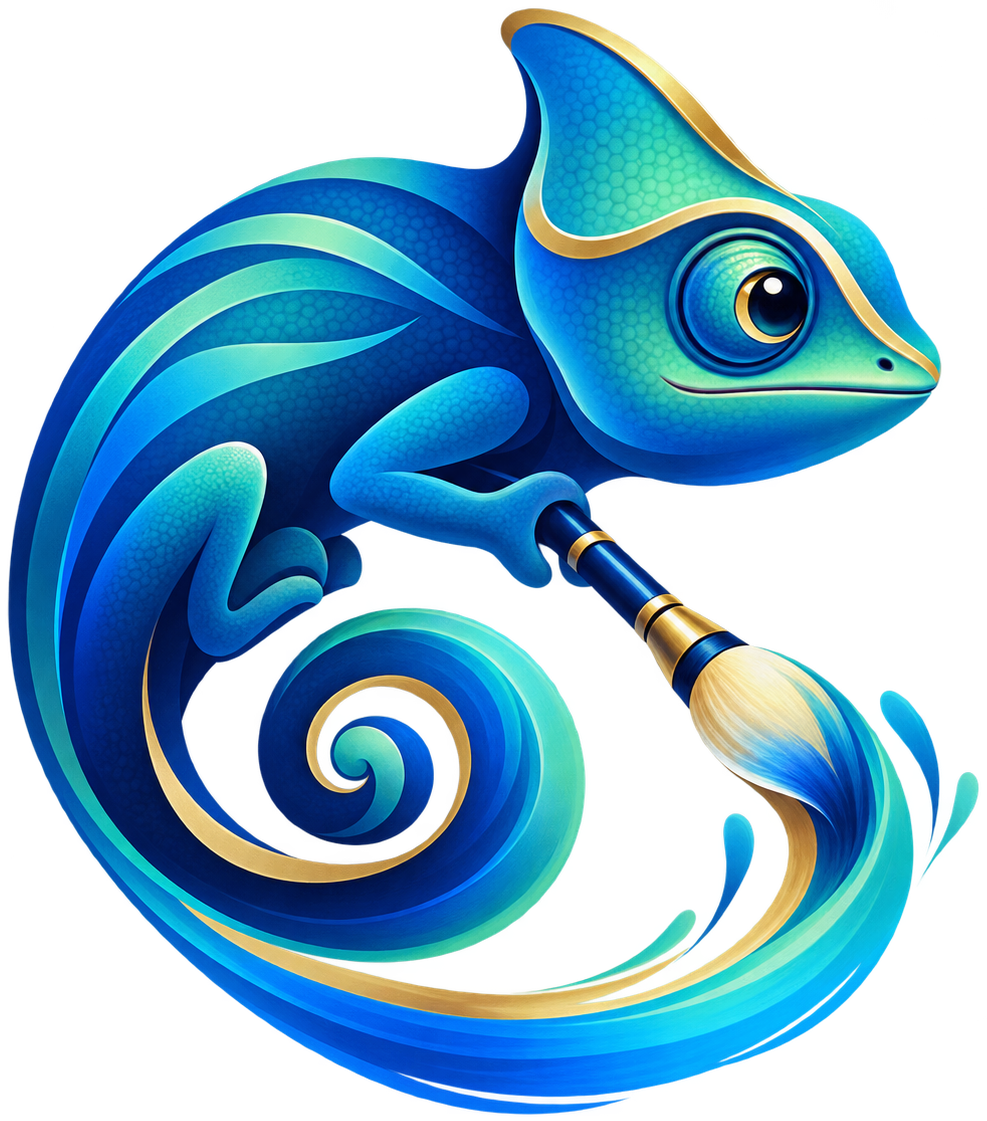}%
  }\hspace{0.25em}%
  MaLiang-Harness: A Programmable Path to \\ Image and Video Generation
}

\vspace{-5pt}
\author{
  \vspace{-25pt}\\
  \textbf{
  Haoyu Zhao$^{1,2,\dagger}$,\quad
  Zihao Zhang$^{1,2}$,\quad
  Xudong Wang$^{1}$,\quad
  Jiaxi Gu$^{3}$,\quad
  }\\[0pt]
  \textbf{
  Zuxuan Wu$^{2}$,\quad
  Yu-Gang Jiang$^{2}$,\quad
  Shuicheng Yan$^{1}$
  }
  \vspace{3pt}\\
  $^1$National University of Singapore
  \quad
  $^2$Fudan University
  \quad
  $^3$Tencent
  \vspace{-4pt}
}

\begin{document}

\maketitle

\begingroup
\renewcommand{\thefootnote}{\fnsymbol{footnote}}

\footnotetext[2]{
Project lead.
}

\endgroup

\begin{abstract}
Executable programs offer explicit control over how images and videos are constructed, but generating runnable code is only the beginning of visual creation. A program can execute correctly while violating the requested composition, appearance, or motion. We define this discrepancy as the Program-to-Visual (P2V) gap and introduce \textsc{MaLiang-Harness}, a unified framework for organizing MLLM-driven visual generation into a persistent process of construction, inspection, and revision. Its central design is to make the evolving visual program, its construction history, and its verification share a common revision reference. We define the Persistent Executable Generation (PEG) state as preserving programs and task context. Traceable Generation Process (TGP) connects edits to rendered evidence, and Revision-aware Editing and Verification (REV) supports restoration and checks the current revision before completion. Together, these mechanisms coordinate planning, execution, and visual feedback across rendering backends. We evaluate 11 powerful closed-source MLLMs on MaLiang-IBench and four on MaLiang-VBench, measuring generation success, visual quality, and computational cost. GPT-6-Astra achieves 100\% generation success on both benchmarks, with 96.0\% of image tasks and 76.9\% of video tasks meeting all quality thresholds. 
The comparison also reveals a mismatch between general capability scores and visual generation performance, with similarly scored models differing substantially in their ability to satisfy visual requirements.
MaLiang-Harness provides a systematic basis for studying how MLLMs translate executable code into visual outcomes, exposing both the potential of programmable generation and the limitations of general benchmarks as predictors of this ability.
The project is available at \href{https://github.com/gulucaptain/MaLiang-Harness}{https://github.com/gulucaptain/MaLiang-Harness}.
\end{abstract}

\section{Introduction}
\label{sec:intro}

\begin{quote}
\centering
\emph{``The idea becomes a machine that makes the art.''}\par
\vspace{1mm}
--- Sol LeWitt
\end{quote}

\begin{figure}[t!]
    \centering
    \includegraphics[width=1.0\linewidth]{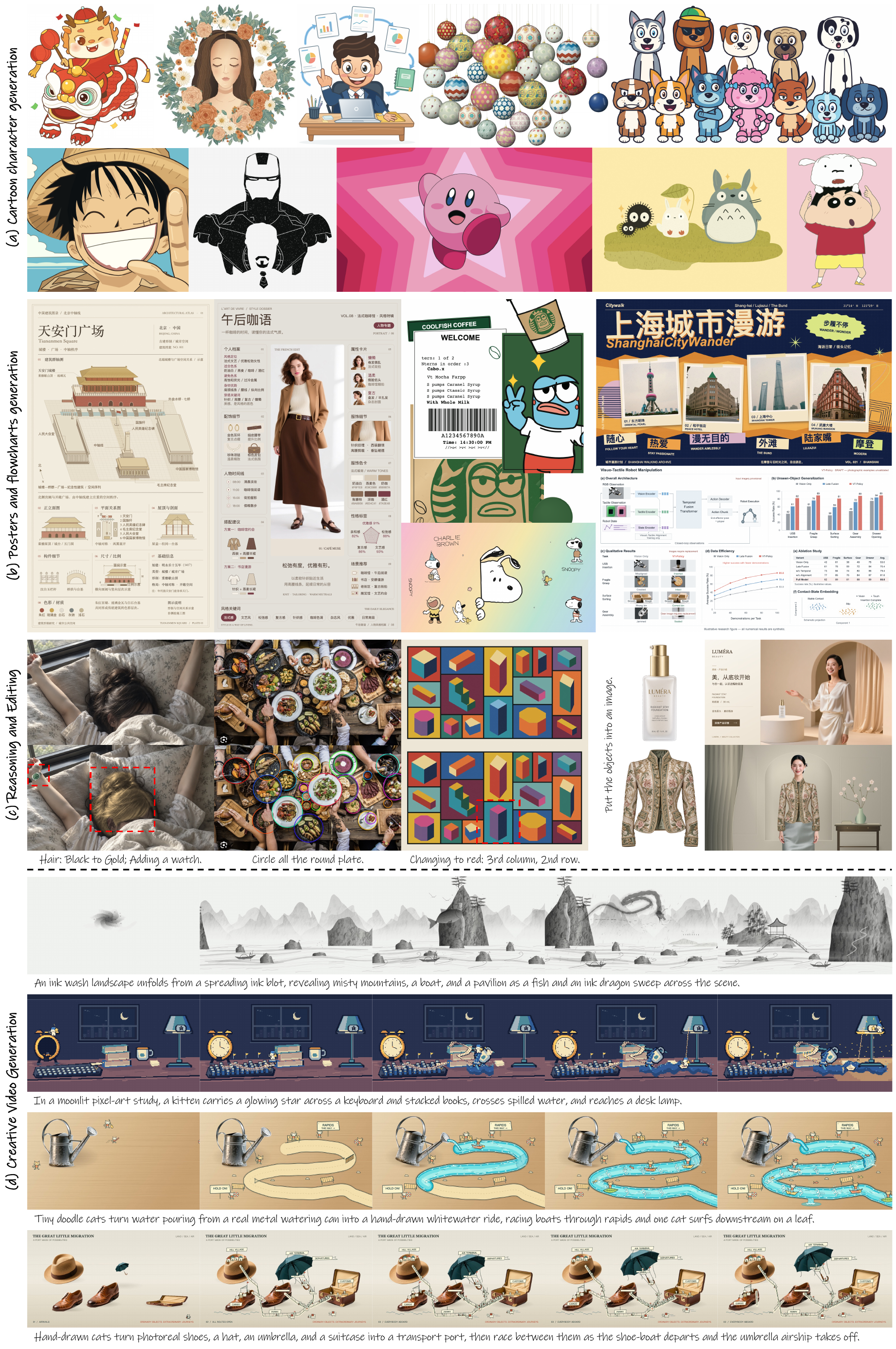}
    \caption{\textbf{MaLiang-Harness aims to explore a programmable path to image and video generation}, enabling diverse image synthesis, visual reasoning and editing, and creative video generation.}
    \label{fig:teaser}
\end{figure}

Over the past decade, advances in high-quality image and video generation have largely followed the paradigm of direct visual synthesis. From generative adversarial networks to diffusion- and flow-based models~\citep{ho2020denoising,lipman2022flow}, these approaches learn the distribution of visual data and directly synthesize pixels or latent visual representations~\citep{rombach2022high}. Although this paradigm has achieved remarkable success, its underlying generation process typically remains implicit. In contrast, the rapidly improving capabilities of multimodal large language models (MLLMs) in semantic understanding, reasoning, and code generation are making a programmable path to image and video generation increasingly viable: \textbf{a model expresses creative intent as an executable visual program, which a renderer subsequently converts into an image or video.}

However, a program can execute correctly yet produce an image or video that fails to satisfy the user's intent. An object may appear in the wrong place, or an animated event may occur at the wrong time. We refer to this discrepancy between program-level correctness and visual requirement satisfaction as the Program-to-Visual (P2V) Gap. Bridging this gap requires the model to reason about the visual consequences of its code. Its initial plan must account for the choice of visual representation and rendering backend. Once the code is executed, the rendered output provides evidence for revising both the program and the plan. This process must preserve the artwork across iterations so that the model can address unmet requirements while retaining access to earlier versions. Each revision also calls for renewed visual assessment because a change can affect requirements that were previously satisfied. These demands motivate a stateful harness that connects creative intent to executable visual representations and supports continued control over the artwork through visual feedback.

\begin{table}[!t]
    \centering
    \caption{\textbf{Comparison of creative-style coverage and generation capabilities.}
    We qualitatively compare MaLiang-Harness with existing image (T2I) and video (T2V) generation models along two dimensions:
    1) creative-style coverage spanning line and flat art, 2D cartoons, pixel art, stylized 3D, mixed media and craft, and photorealism;
    and 2) generation capabilities including controllability over visual content and spatiotemporal structure, editability of the generated artwork, and traceability of the creation process.
    MaLiang-Harness supports diverse visual styles through executable visual programs, while providing explicit control, targeted editing, and an inspectable construction process.
    {\scriptsize \MLfull}~Supported. 
    {\scriptsize \MLpartial}~Partially supported. 
    {\scriptsize \MLempty}~Not established.}
    \label{tab:style_coverage}
    
    \small
    \setlength{\tabcolsep}{5pt}
    \renewcommand{\arraystretch}{1.35}

    \resizebox{\linewidth}{!}{%
    \begin{tabular}{@{}ll|cccccc|ccc@{}}
        \toprule
        
        \textbf{Type} & \textbf{Method}
        & \stylehead{pencil-alt}{Line\&Flat \\Art}
        & \stylehead{cat}{2D\\Cartoon}
        & \stylehead{th-large}{Pixel\\Art}
        & \stylehead{cube}{Stylized\\3D}
        & \stylehead{layer-group}{Mixed\\ \& Craft}
        & \stylehead{camera}{Photo-\\realistic}
        & \stylehead{sliders-h}{Contro-\\llability}
        & \stylehead{edit}{Edita-\\bility}
        & \stylehead{search}{Process\\Traceability}
        \\
        
        \midrule
        
        \multirow{2}{*}{\textbf{Image}}
        & T2I
        & \MLfull & \MLfull & \MLfull
        & \MLfull & \MLfull & \MLfull
        & \MLfull & \MLfull & \MLempty
        \\
        
        & \textbf{Ours}
        & \MLfull & \MLfull & \MLfull
        & \MLfull & \MLfull & \MLpartial
        & \MLfull & \MLfull & \MLfull
        \\
        
        \midrule

        \multirow{2}{*}{\textbf{Video}}
        & T2V
        & \MLfull & \MLfull & \MLfull
        & \MLfull & \MLfull & \MLfull
        & \MLfull & \MLfull & \MLempty
        \\
        
        & \textbf{Ours}
        & \MLfull & \MLfull & \MLfull
        & \MLfull & \MLfull & \MLpartial
        & \MLfull & \MLfull & \MLfull
        \\

        \bottomrule
    \end{tabular}%
    }
\end{table}

To address this challenge, we introduce \raisebox{-0.15em}{\includegraphics[height=1em]{figures/maliang2.png}}\,\textbf{MaLiang-Harness}, a unified interface connecting creative intent to multiple executable visual representations.
It unifies programmable image and video generation through a shared protocol for state management, rendering, and revision-aware verification.
Through this interface, an MLLM selects suitable rendering backends and exercises explicit control over the artwork through visual feedback. Code directly specifies how visual content is constructed, from spatial composition to motion over time. Optional image assets complement this programmatic structure when rich appearance is difficult to express through code alone. Images and videos share the same creation framework: at a fixed program revision, an image is rendered at a specified content time, while a video samples the temporal behavior encoded by the program. Table~\ref{tab:style_coverage} summarizes the creative-style coverage and functional capabilities.

Furthermore, MaLiang-Harness organizes creation into a loop in which the MLLM plans and generates code, then uses execution results and visual feedback to guide revision. Three designs support this process:
\textbf{1)} \textit{Persistent Executable Generation (PEG) State} maintains the executable definition of the evolving artwork across iterations. It brings programs and assets together with their spatiotemporal organization, task requirements, the current plan, and revision index. This gives the model a stable object to inspect and modify as creation progresses.
\textbf{2)} \textit{Traceable Generation Process (TGP)} makes the construction of the artwork inspectable through explicit programs and recorded operations. Intermediate renderings expose the visual effects of these operations, helping the model and the user relate an observed discrepancy to the components or changes that may have produced it.
\textbf{3)} \textit{Revision-aware Editing and Verification (REV)} allows the model to restore earlier artwork content as a new revision while retaining current task requirements. Observations and verification results are associated with specific revisions, and every committed update requires renewed assessment before completion.
We demonstrate that these designs enable the MLLM to translate visual feedback into targeted revisions of the evolving artwork, helping bridge the P2V gap.

\begin{figure}[t!]
    \centering
    \includegraphics[width=1.0\linewidth]{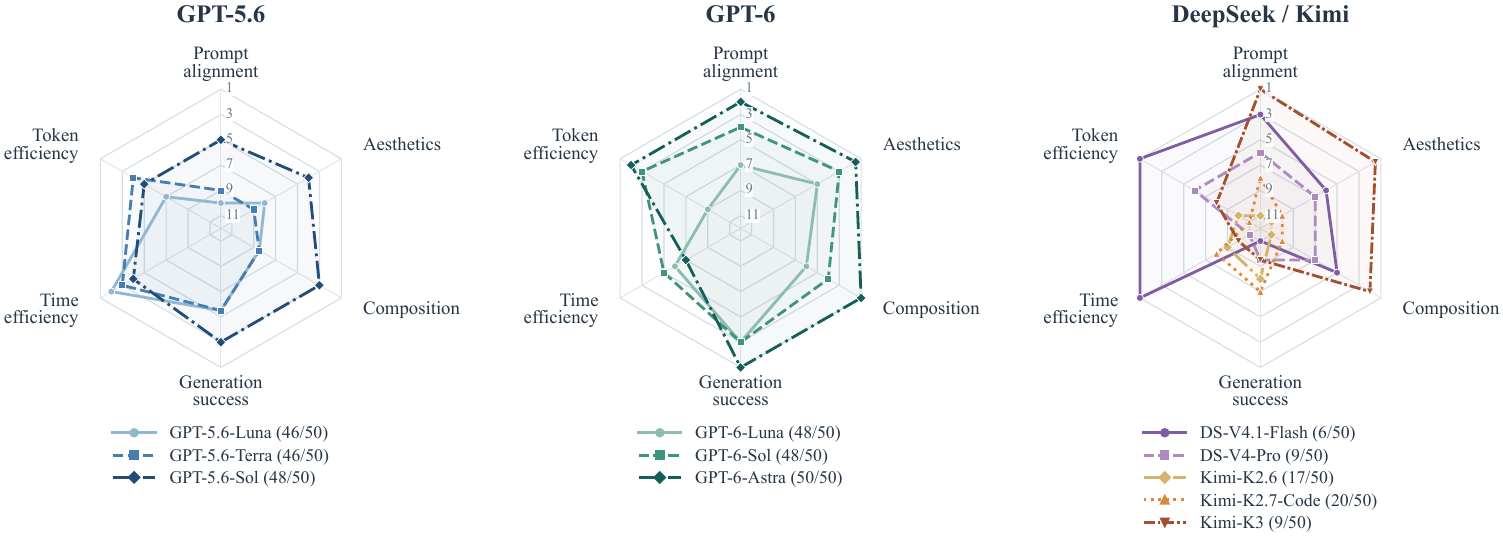}
    \caption{\textbf{Model rankings on MaLiang-IBench.} Ranks are computed across 11 models; better ranks lie farther from the center, with ties averaged. GPT models achieve 92--100\% generation success, versus 12--40\% for DeepSeek and Kimi. Mean quality scores use successful outputs (counts in parentheses). Cost ranks favor less time and fewer tokens.}
    \label{fig:rank_radar}
\end{figure}

We evaluate 11 MLLMs on MaLiang-IBench and four on MaLiang-VBench. GPT-6-Astra~\citep{gpt6} achieves a 100\% generation success rate on both benchmarks. Its outputs meet every quality threshold on 96.0\% of image tasks and 76.9\% of video tasks, compared with 86.0\% and 38.5\% for GPT-5.6-Sol. These percentages use the full task set of each benchmark as the denominator. These results distinguish successful execution from satisfaction of visual requirements.
Fig.~\ref{fig:rank_radar} compares 11 models on MaLiang-IBench across visual quality, generation success, and computational cost, revealing different model rankings across these dimensions.
Fig.~\ref{fig:teaser} complements these comparisons with examples of image generation, visual understanding and editing, and video generation.

Our main contributions are:
\begin{itemize}
    \item We introduce and define the \textbf{Program-to-Visual (P2V) gap} as the discrepancy between program-level correctness and visual requirement satisfaction, motivating a stateful formulation of visual program generation through construction, inspection, and revision.

    \item We introduce \textbf{MaLiang-Harness}, a unified framework for programmable image and video generation. Persistent Executable Generation state, a Traceable Generation Process, and Revision-aware Editing and Verification support continued refinement, inspection of construction histories, and verification of the current output.

    \item We evaluate 11 MLLMs on MaLiang-IBench and four on MaLiang-VBench, characterizing differences in generation success, visual quality, and computational cost. Comparison with public general-capability scores shows that similar benchmark performance can correspond to substantially different visual generation outcomes.
\end{itemize}

\section{Related Work}
\label{sec:related}

\paragraph{Image and Video Generation.}
Diffusion models synthesize images through learned denoising processes~\citep{ho2020denoising}, with latent diffusion reducing the cost of high-resolution synthesis~\citep{rombach2022high}. Flow matching provides a related framework for learning continuous generative dynamics~\citep{lipman2022flow}. Video Diffusion Models and Stable Video Diffusion extend learned visual synthesis to temporal content~\citep{ho2022video,blattmann2023stable}. These approaches also support explicit conditioning: ControlNet, for example, introduces spatial guidance through inputs such as edges and depth~\citep{zhang2023adding}. Our focus is on the representation through which visual content is constructed and revised. MaLiang-Harness maintains an executable artwork whose geometry and temporal behavior can be inspected and edited.

\paragraph{Executable Visual Representations.}
Prior work establishes several ways to connect language models with executable visual representations. VISPROG generates modular programs for visual reasoning and image editing, exposing intermediate results as inspectable rationales~\citep{gupta2023visual}. Design2Code evaluates the translation of webpage screenshots into renderable implementations~\citep{si2025design2code}. In three-dimensional graphics, BlenderAlchemy combines a vision-based edit generator with a state evaluator to search for edits that realize a user's design intent~\citep{huang2024blenderalchemy}. These studies demonstrate that programs can mediate visual understanding and construction. MaLiang-Harness builds on this premise by organizing image and video creation around a persistent artwork state across several rendering backends.

\paragraph{Agent Harnesses.}
Code as Agent Harness provides a broad account of code as infrastructure for reasoning, action, and stateful execution~\citep{ning2026code}. Show-Harness demonstrates how a semantic interface can connect VLM decisions to executable actions across robot embodiments~\citep{chen2026show}. Within visual generation, OmniHarness learns reusable symbolic policies from verified executions and uses intermediate feedback for refinement and recovery~\citep{xu2026omniharness}.
MaLiang-Harness manages executable artwork through interfaces and feedback, linking visual construction to its edit history and revision-specific evidence.

\section{MaLiang-Harness}
\label{sec:method}

\begin{figure}[t!]
    \centering
    \includegraphics[width=1.0\linewidth]{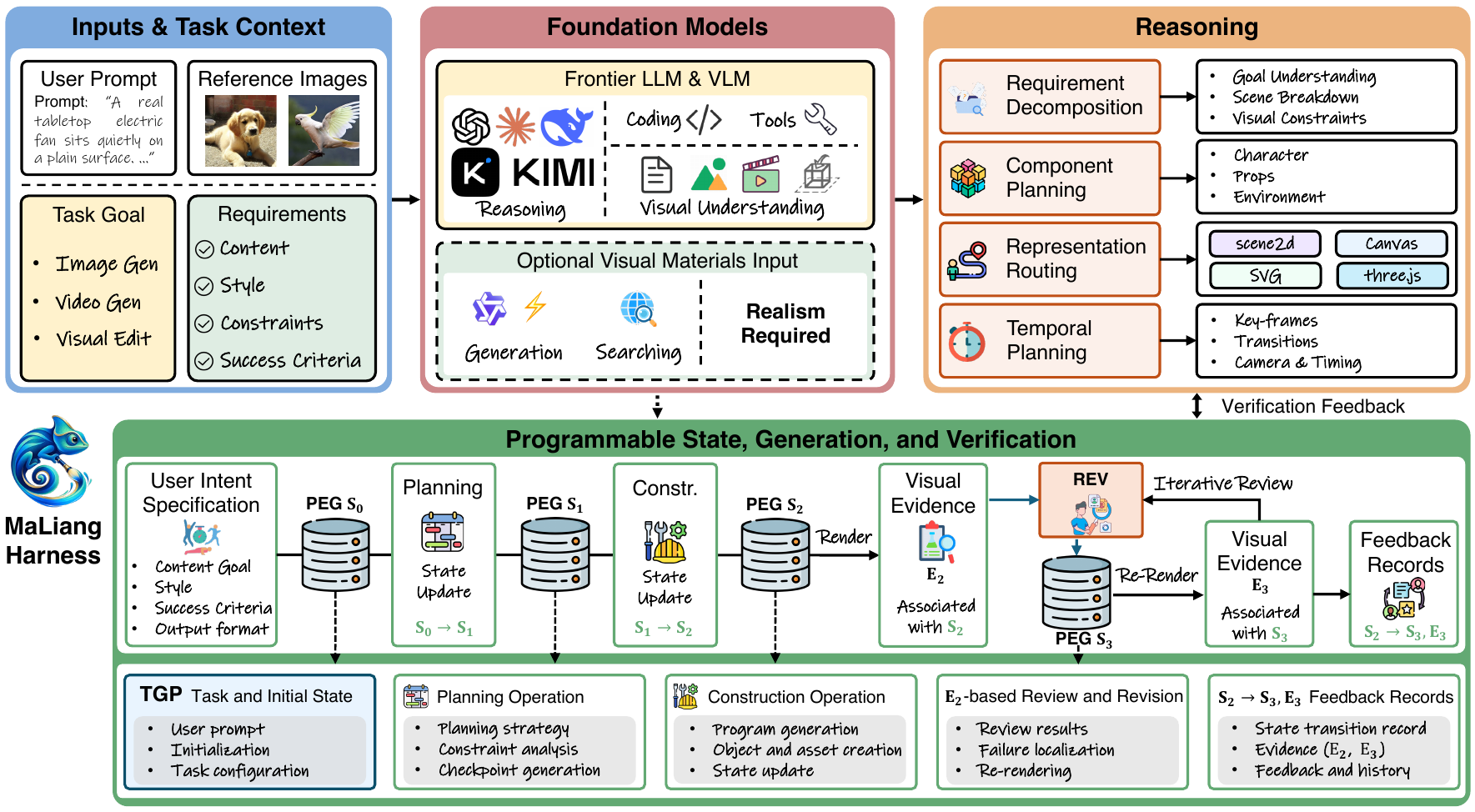}
    \caption{\textbf{Overview of the MaLiang-Harness for image and video generation.} Persistent Executable Generation (PEG) state maintains the program and task context across revisions. Traceable Generation Process (TGP) links generation operations to state changes and visual results. Revision-aware Editing and Verification (REV) supports revision from historical states and assesses the current output using evidence from the same revision.}
    \label{fig:framework_overview}
\end{figure}

MaLiang-Harness explores an alternative path to image and video generation in which an MLLM translates a prompt $p$ into executable visual programs through a unified generation interface, as shown in Fig.~\ref{fig:framework_overview}. The framework uses MLLM planning and code generation to construct expressive visual compositions that renderers turn into images and videos, without requiring a diffusion or flow-matching process. Visual feedback guides iterative refinement of the generation under explicit spatial and temporal control.
The carefully designed persistent executable generation state, traceable generation process, and revision-aware editing and verification associate each operation with the states it connects and each visual assessment with the revision that produced its evidence.

\subsection{Unified Programmatic Visual Generation Interface}
\label{sec:unified_creation_interface}

MaLiang-Harness provides a shared interaction protocol across rendering backends (\textit{e.g.,} Canvas, SVG, Scene2d, or Three.js) that supports state inspection and program or asset editing alongside rendering and requirement review.
State updates identify the revision they modify, while execution results and rendered observations are linked to the corresponding revisions. Each backend retains its native executable representation, allowing the harness to express visual content through drawing and animation code within a common generation process.

Given a prompt $p$ and output specification $\omega$, MaLiang-Harness plans the appearance, spatial composition, and temporal dynamics, selects an executable representation and compatible backend $b$, and generates drawing and animation code. The program may incorporate user-provided assets or, when the desired appearance is difficult to construct through code, assets obtained through image generation or search.
The backend renders the resulting representation into an image or video that the MLLM evaluates against the task requirements, grounding subsequent code revisions in observed visual discrepancies rather than execution success alone.

\noindent \textbf{Persistent Executable Generation (PEG) State.}
The harness maintains a persistent executable generation state that provides a common reference for planning, execution, and revision.
At revision $k$, we define:
\begin{equation}
    S_k = \bigl(P_k, A_k, Z_k, C_k, k\bigr),
    \label{eq:persistent-artwork-state}
\end{equation}
where $P_k$ denotes the visual program together with its backend identifier, and $A_k$ contains any associated assets. 
Program and asset contents are retained with content hashes, while $\omega$ specifies the output dimensions and format together with the rendering seed, and any applicable video timing parameters.
$Z_k$ describes the spatial composition and temporal dynamics through explicit scene attributes or definitions embedded in $P_k$, without requiring a separately maintained scene representation.
The generation context $C_k$ retains $p$, $\omega$, user-supplied requirements, and the current generation plan. Planning updates may append requirements but cannot remove or weaken existing ones. Changes to inherited requirements in a separate editing task must cite the user's new instruction.

\textit{Initial state.} The initial state $S_0$ contains $p$ and $\omega$ and may have no executable content, which is refined through subsequent edits.
We distinguish updates to this state from the visual rendering:
\begin{equation}
    S_{k+1} = \mathcal{E}(S_k, a_k),
    \qquad
    I_k(t) = \mathcal{R}_b(S_k, t; \omega),
    \label{eq:creation-interface}
\end{equation}
where $\mathcal{E}$ commits an edit $a_k$ after validating its inputs and checking that its source revision is current.
Each commit, including restoration of historical content, receives the next unused revision index within the run.
The harness retains the preceding snapshot so that subsequent generation can revisit an earlier version rather than overwrite its executable definition.

\noindent
\begin{minipage}[t]{0.38\textwidth}
\vspace{0pt}
\centering
\includegraphics[width=0.8\linewidth]{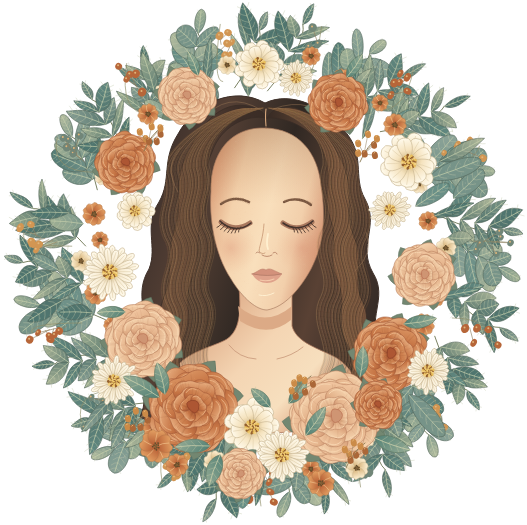}
\par\medskip
\captionof{figure}{\textbf{Example of code-driven image synthesis.}
The ``botanical portrait'' is rendered from the visual program $P_k$ of a PEG state without image assets. Algorithm~\ref{alg:botanical} summarizes how the program specifies geometry, appearance, and spatial composition through Canvas drawing operations.}
\label{fig:peg-botanical}
\end{minipage}\hfill
\begin{minipage}[t]{0.57\textwidth}
\vspace{0pt}
\captionof{algorithm}{Generated Canvas program for T2I.}
\label{alg:botanical}
\begin{algorithmic}[1]
\Require Canvas $1024\times576$, seed $s$
\State $\mathbf{c}\gets(512,288)$
\State $R_\ell\gets\mathrm{PRNG}(\mathrm{hash}(s,\ell))$
       for each layer $\ell$
\State Fill the canvas with white.

\For{$i=0,\ldots,25$}
  \State $\theta_i\gets 2\pi i/26+\epsilon_i$,
         \quad $r_i\gets167+18u_i$
  \State $\mathbf{p}_i\gets
         \mathbf{c}+r_i(\cos\theta_i,\sin\theta_i)$
  \State Draw a curved branch at $\mathbf{p}_i$;
         add leaves and veins.
\EndFor

\State Draw Bézier paths for hair, face, and shoulders.
\State Shade the face; add closed eyes and hair strands.

\For{each rose at a predefined position}
  \For{$k=0,\ldots,4$}
    \State $n_k\gets[8,7,6,5,4]_k$
    \For{$j=0,\ldots,n_k-1$}
      \State $\phi_{k,j}\gets2\pi j/n_k+\delta_{k,j}$
      \State Draw a rotated Bézier petal at
             angle $\phi_{k,j}$.
    \EndFor
  \EndFor
\EndFor
\State Add daisies, small blossoms, and texture.
\State Generation finished.
\end{algorithmic}
\vspace{10pt}
\end{minipage}

\textit{Renderable state.} For a renderable state, $\mathcal{R}_b$ evaluates the visual program at content time $t$ under $\omega$ using the selected backend $b$. An image corresponds to evaluation at a specified time, whereas a video is obtained by sampling the temporal evolution encoded in the same representation. The index $k$ therefore tracks revisions of the generation state rather than progression through the generated content; an update to the generation plan can advance $k$ without changing the rendered output.
Fig.~\ref{fig:peg-botanical} pairs a visual program with its rendered output at a fixed PEG revision, showing how the executable representation specifies the resulting image composition.

The harness records each committed edit with its source and resulting revisions, linking the operation history to the evolution of the executable state. Rendered observations and verification results are associated with the revision they assess, but collecting this evidence does not itself create a new PEG revision. The MLLM uses the revision-specific evidence together with the requirements in $C_k$ to determine subsequent edits.
The generation state thus provides a shared revision reference for the construction history tracked by TGP and the visual assessments maintained by REV.

\subsection{Traceable Generation Process}
\label{sec:tcp}

While the PEG state preserves the executable visual representation at each revision, the traceable generation process (TGP) records how this representation is constructed and modified through generation operations.
For the $j$-th recorded operation, the harness maintains:
\begin{equation}
    \tau_j =
    \bigl(o_j, x_j, y_j, k_j^{-}, k_j^{+}\bigr),
    \label{eq:generation-trace}
\end{equation}
where $o_j$ denotes the executed operation, $x_j$ and $y_j$ are its input arguments and execution result (including returned errors), and $k_j^{-}$ and $k_j^{+}$ identify the source and resulting PEG revisions.
The operation index $j$ is distinct from the revision index $k$, since operations that do not commit a state update, including rendering and inspection, retain the same revision index.
Comparing the corresponding snapshots exposes changes to the visual program and its associated state without requiring a common scene representation across rendering backends.

The harness links each rendered observation to the PEG revision evaluated by the renderer and retains the actual output, sampled timestamps, and any crop specification. These records identify the visual evidence used in a review even when custom code introduces nondeterminism beyond the configured seed.
For example, an edit to a drawing function or motion trajectory can be located in the operation history and compared with renderings of the relevant revisions to investigate discrepancies in spatial composition or temporal dynamics.
Rendering and inspection need not follow every edit, so an executed modification and a visually inspected revision remain distinguishable in the trace.

TGP supports diagnosis of the P2V gap by connecting recorded program changes with revision-specific visual evidence, rather than treating successful execution as confirmation of the intended visual result.
Its traceability concerns explicit generation operations and their execution results, not the MLLM's internal reasoning.
These operation-to-revision and evidence-to-revision associations provide the basis for subsequent inspection and revision, while the validity of visual assessments after further edits is addressed by REV.

\subsection{Revision-aware Editing and Verification}
\label{sec:rev}

\begin{figure}[ht]
    \centering
    \includegraphics[width=0.9\linewidth]{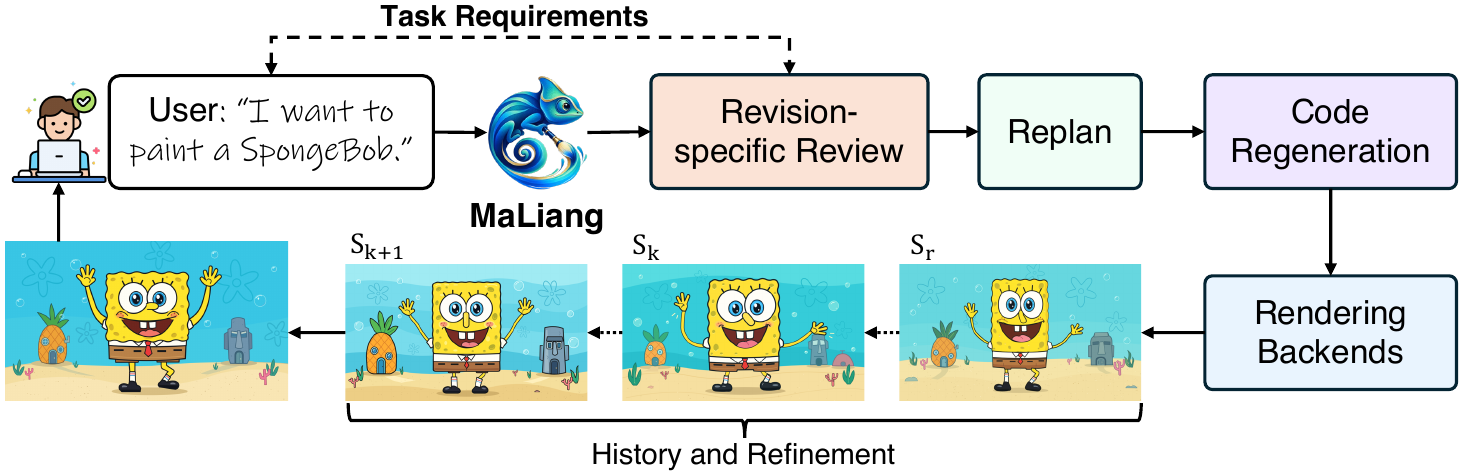}
    \caption{\textbf{Revision-aware editing and verification.} Edits produce new PEG revisions, and visual evidence is associated with the revision it assesses. Historical evidence supports comparison; delivery requires verification of the current revision.}
    \label{fig:revision_refinement}
\end{figure}

As shown in Fig.~\ref{fig:revision_refinement}, revision-aware editing and verification (REV) closes the generation loop by verifying edits to PEG states using the revision-linked evidence recorded by TGP.
Within a run, restoring historical content from $S_r$ ($r\leq k$) commits a new state $S_{k+1}$ while preserving current task requirements; the trace records the restored revision $r$. A separate editing task initializes its own revision sequence from the selected snapshot, retains a link to the source run and revision, and incorporates the new user instruction into $C_0$. Subsequent edits remain traceable through TGP.

For each visual requirement $h_i$ in $C_k$, REV maintains a review $q_i=(k_i,E_i,v_i)$, where $E_i$ contains requirement-specific visual evidence from revision $k_i$ and $v_i\in\{\mathrm{pass},\mathrm{fail},\mathrm{uncertain}\}$. Full-frame renderings or spatial crops support appearance and composition checks, while temporal requirements use ordered samples spanning the specified interval at three or more distinct timestamps.
Such samples support temporal assessment but do not establish continuity between frames. A review applies to the current state only when $k_i=k$.
After each commit, even if only the plan changes, the current revision must be reviewed again and its checkpoints reapproved. Recording observations and reviews leaves $k$ unchanged. Moreover, the current revision is ready for delivery only when the following conditions hold:
\begin{equation}
\begin{aligned}
\operatorname{Ready}(S_k)
={}& \operatorname{ExportOK}(S_k)
\land \operatorname{CheckpointOK}(k) \\
&{}\land
\bigwedge_{h_i\in\mathcal{H}_k}
\left[
k_i=k \land E_i\neq\varnothing
\land v_i=\mathrm{pass}
\right],
\end{aligned}
\label{eq:revision-verification}
\end{equation}
where $\mathcal{H}_k$ contains the visual and temporal requirements marked as mandatory for completion in $C_k$. 
$\operatorname{ExportOK}$ verifies that source files and assets exist and match their recorded hashes. It also checks applicable object and event constraints and confirms that the current revision's export conforms to $\omega$. These export checks cover format, dimensions, and video timing. During planning, the MLLM organizes all mandatory requirements into checkpoints. $\operatorname{CheckpointOK}$ holds when every checkpoint passes the required checks and reviews for the current revision. Missing evidence or reviews marked as failed or uncertain require further inspection or revision while budget remains. This criterion ties visual verification to the delivered revision. MLLM verdicts remain self-assessments rather than independent measures of perceptual quality.

\section{Experiments} \label{sec:experiments}
\subsection{Experimental Setup}
\label{sec:data_sources_metrics}

\paragraph{Tasks and models.}
We evaluate MaLiang-Harness on \textbf{MaLiang-IBench} with 50 text-to-image prompts and \textbf{MaLiang-VBench} with 13 text-to-video prompts. Both sets cover diverse visual styles.
We evaluate 11 models from the DeepSeek~\citep{xu2026deepseek}, Kimi~\citep{team2026kimi}, and GPT~\citep{gpt5,gpt6} families on MaLiang-IBench and four models on MaLiang-VBench.
Each model constructs and refines executable visual programs through MaLiang-Harness.

\paragraph{Evaluation metrics.}
Our evaluation covers two aspects: computational cost and generation quality. We report generation success, failures, and the subset of failures caused by token-budget exhaustion (Token Limit). Success requires a decodable output that passes the harness's completion checks. \textbf{(1) Computational cost.} We report generation time, model calls, and token usage. Cumulative time includes failed attempts and retries. \emph{Time/Qualified} normalizes this time by the number of images meeting all quality criteria; \emph{Time/Success} uses the number of successfully generated videos. Note that DeepSeek-V4-Pro is evaluated without visual feedback. \textbf{(2) Generation quality.} We primarily employ GPT-6-Sol as a judge. It evaluates image quality on a five-point scale for prompt alignment, aesthetics, and composition. Video evaluation includes motion coherence as an additional criterion and uses 12 temporally ordered frames from each successful video. Mean image-quality scores are computed over successful generations only. Per-dimension counts report the number of samples scoring at least 4, while \emph{All Criteria} counts samples meeting all three image criteria or all four video criteria. Note that failed generations are excluded from all quality-threshold counts.

\subsection{Comparisons}
\label{sec:results}

\begin{table*}[t]
    \centering
    \definecolor{TtwoBlue}{RGB}{183,203,242}
    \definecolor{TtwoOrange}{RGB}{248,215,199}    \def\TtwoShade#1#2{\begingroup\setlength{\fboxsep}{2pt}\colorbox{#1}{\makebox[3.2em]{#2}}\endgroup}
    \caption{\textbf{Generation success and computational cost on MaLiang-IBench.} ``Token Limit'' denotes the subset of failures in which a run exhausts its allocated token budget, potentially due to reasoning stagnation and unbounded agentic loops. ``Orange/blue'' indicates ``higher/lower'' values relative to DeepSeek-V4.1-Flash, which do not imply ``better/worse'' performance given differing success rates.}
    \label{tab:llm_image_benchmark}
    \resizebox{\textwidth}{!}{%
        \begin{tabular}{lccccccccc}
            \toprule
            & \multicolumn{3}{c}{\textbf{Task Completion}}
            & \multicolumn{3}{c}{\textbf{Generation Time}}
            & \multicolumn{3}{c}{\textbf{Model Usage}} \\
            \cmidrule(lr){2-4}
            \cmidrule(lr){5-7}
            \cmidrule(lr){8-10}
            \textbf{Model}
            & Success $\uparrow$
            & Failures $\downarrow$
            & Token Limit $\downarrow$
            & Total Time
            & Time/Qualified $\downarrow$
            & Time/Task
            & Calls
            & Input Tokens
            & Output Tokens \\
            & (\%) & (cases) & (cases) & (task-h) & (min/image)
            & (min/case) & (calls/case) & (k/case) & (k/case) \\
            \midrule
            DeepSeek-V4.1-Flash & \TtwoShade{white}{12.0} & \TtwoShade{white}{44} & \TtwoShade{white}{5} & \TtwoShade{white}{2.71} & \TtwoShade{white}{27.15} & \TtwoShade{white}{1.91} & \TtwoShade{white}{2.48} & \TtwoShade{white}{100.8} & \TtwoShade{white}{24.2} \\
            DeepSeek-V4-Pro & \TtwoShade{TtwoOrange!71}{18.0} & \TtwoShade{TtwoBlue!26}{41} & \TtwoShade{TtwoBlue!63}{3} & \TtwoShade{TtwoOrange!100}{6.42} & \TtwoShade{TtwoOrange!88}{48.12} & \TtwoShade{TtwoOrange!100}{7.36} & \TtwoShade{TtwoOrange!90}{4.50} & \TtwoShade{TtwoOrange!99}{198.9} & \TtwoShade{TtwoOrange!80}{39.6} \\
            \midrule
            Kimi-K2.6 & \TtwoShade{TtwoOrange!100}{34.0} & \TtwoShade{TtwoBlue!50}{33} & \TtwoShade{TtwoBlue!89}{1} & \TtwoShade{TtwoOrange!100}{6.27} & \TtwoShade{TtwoOrange!100}{62.69} & \TtwoShade{TtwoOrange!100}{5.14} & \TtwoShade{TtwoOrange!100}{9.14} & \TtwoShade{TtwoOrange!100}{660.0} & \TtwoShade{TtwoOrange!100}{48.2} \\
            Kimi-K2.7-Code & \TtwoShade{TtwoOrange!100}{40.0} & \TtwoShade{TtwoBlue!56}{30} & \TtwoShade{TtwoBlue!89}{1} & \TtwoShade{TtwoOrange!100}{5.97} & \TtwoShade{TtwoOrange!31}{29.84} & \TtwoShade{TtwoOrange!94}{3.59} & \TtwoShade{TtwoOrange!100}{12.94} & \TtwoShade{TtwoOrange!100}{1056.5} & \TtwoShade{TtwoOrange!100}{59.0} \\
            Kimi-K3 & \TtwoShade{TtwoOrange!71}{18.0} & \TtwoShade{TtwoBlue!26}{41} & \TtwoShade{TtwoBlue!100}{0} & \TtwoShade{TtwoOrange!96}{5.23} & \TtwoShade{TtwoOrange!53}{34.88} & \TtwoShade{TtwoOrange!100}{5.89} & \TtwoShade{TtwoOrange!37}{2.82} & \TtwoShade{TtwoOrange!100}{308.9} & \TtwoShade{TtwoBlue!51}{17.8} \\
            \midrule
            GPT-5.6-Luna & \TtwoShade{TtwoOrange!100}{92.0} & \TtwoShade{TtwoBlue!95}{4} & \TtwoShade{TtwoBlue!100}{0} & \TtwoShade{TtwoBlue!63}{1.65} & \TtwoShade{TtwoBlue!91}{4.49} & \TtwoShade{TtwoOrange!19}{1.98} & \TtwoShade{TtwoOrange!100}{11.10} & \TtwoShade{TtwoOrange!100}{271.6} & \TtwoShade{TtwoBlue!78}{9.3} \\
            GPT-5.6-Terra & \TtwoShade{TtwoOrange!100}{92.0} & \TtwoShade{TtwoBlue!95}{4} & \TtwoShade{TtwoBlue!100}{0} & \TtwoShade{TtwoBlue!60}{1.73} & \TtwoShade{TtwoBlue!92}{4.32} & \TtwoShade{TtwoOrange!29}{2.07} & \TtwoShade{TtwoOrange!100}{8.32} & \TtwoShade{TtwoOrange!92}{186.0} & \TtwoShade{TtwoBlue!82}{8.0} \\
            GPT-5.6-Sol & \TtwoShade{TtwoOrange!100}{96.0} & \TtwoShade{TtwoBlue!98}{2} & \TtwoShade{TtwoBlue!89}{1} & \TtwoShade{TtwoBlue!36}{2.35} & \TtwoShade{TtwoBlue!94}{3.28} & \TtwoShade{TtwoOrange!69}{2.82} & \TtwoShade{TtwoOrange!100}{6.98} & \TtwoShade{TtwoOrange!100}{209.2} & \TtwoShade{TtwoBlue!68}{13.1} \\
            \midrule
            GPT-6-Luna & \TtwoShade{TtwoOrange!100}{96.0} & \TtwoShade{TtwoBlue!98}{2} & \TtwoShade{TtwoBlue!100}{0} & \TtwoShade{TtwoOrange!58}{3.61} & \TtwoShade{TtwoBlue!90}{4.92} & \TtwoShade{TtwoOrange!84}{3.27} & \TtwoShade{TtwoOrange!100}{14.70} & \TtwoShade{TtwoOrange!100}{459.2} & \TtwoShade{TtwoBlue!57}{16.4} \\
            GPT-6-Sol & \TtwoShade{TtwoOrange!100}{96.0} & \TtwoShade{TtwoBlue!98}{2} & \TtwoShade{TtwoBlue!100}{0} & \TtwoShade{TtwoBlue!6}{2.70} & \TtwoShade{TtwoBlue!93}{3.52} & \TtwoShade{TtwoOrange!83}{3.24} & \TtwoShade{TtwoOrange!100}{7.42} & \TtwoShade{TtwoOrange!89}{180.1} & \TtwoShade{TtwoBlue!76}{10.2} \\
            GPT-6-Astra & \TtwoShade{TtwoOrange!100}{100.0} & \TtwoShade{TtwoBlue!100}{0} & \TtwoShade{TtwoBlue!100}{0} & \TtwoShade{TtwoOrange!30}{2.96} & \TtwoShade{TtwoBlue!93}{3.70} & \TtwoShade{TtwoOrange!93}{3.55} & \TtwoShade{TtwoOrange!100}{6.34} & \TtwoShade{TtwoOrange!64}{142.0} & \TtwoShade{TtwoBlue!82}{8.1} \\
            \bottomrule
        \end{tabular}%
    }
\end{table*}

\begin{table*}[t]
    \centering
    \definecolor{TthreeBlue}{RGB}{183,203,242}
    \definecolor{TthreeOrange}{RGB}{248,215,199}
    \definecolor{TthreeGray}{RGB}{205,205,205}    \def\TthreeShade#1#2{\begingroup\setlength{\fboxsep}{2pt}\colorbox{#1}{\makebox[3.2em]{#2}}\endgroup}
    \caption{\textbf{Visual quality on MaLiang-IBench.} Quality counts include samples scoring at least 4 on a five-point scale. Means cover successful samples only. Gray cells mark DeepSeek and Kimi results, whose low success rates limit the comparability of mean quality scores. Bold marks the best GPT results. ``Blue/orange'' denotes ``higher/lower'' values relative to GPT-5.6-Luna.}
    \label{tab:llm_image_delivery_quality}
    \resizebox{\textwidth}{!}{%
        \begin{tabular}{lcccccccc}
            \toprule
            & \multicolumn{2}{c}{\textbf{Generation Outcomes}}
            & \multicolumn{3}{c}{\textbf{Quality Threshold Counts}}
            & \multicolumn{3}{c}{\textbf{Mean Quality Scores}} \\
            \cmidrule(lr){2-3}
            \cmidrule(lr){4-6}
            \cmidrule(lr){7-9}
            \textbf{Model}
            & Successful $\uparrow$
            & All Criteria $\uparrow$
            & Alignment $\uparrow$
            & Aesthetics $\uparrow$
            & Composition $\uparrow$
            & Alignment $\uparrow$
            & Aesthetics $\uparrow$
            & Composition $\uparrow$ \\
            & (cases) & (cases) & (cases) & (cases) & (cases)
            & (1--5) & (1--5) & (1--5) \\
            \midrule
            DeepSeek-V4.1-Flash & \TthreeShade{TthreeGray!20}{6} & \TthreeShade{TthreeGray!20}{6} & \TthreeShade{TthreeGray!20}{6} & \TthreeShade{TthreeGray!20}{6} & \TthreeShade{TthreeGray!20}{6} & \TthreeShade{TthreeGray!83}{4.33} & \TthreeShade{TthreeGray!78}{4.00} & \TthreeShade{TthreeGray!81}{4.17} \\
            DeepSeek-V4-Pro & \TthreeShade{TthreeGray!37}{9} & \TthreeShade{TthreeGray!47}{8} & \TthreeShade{TthreeGray!43}{8} & \TthreeShade{TthreeGray!43}{8} & \TthreeShade{TthreeGray!54}{9} & \TthreeShade{TthreeGray!67}{4.11} & \TthreeShade{TthreeGray!67}{3.89} & \TthreeShade{TthreeGray!74}{4.11} \\
            \midrule
            Kimi-K2.6 & \TthreeShade{TthreeGray!83}{17} & \TthreeShade{TthreeGray!20}{6} & \TthreeShade{TthreeGray!31}{7} & \TthreeShade{TthreeGray!43}{8} & \TthreeShade{TthreeGray!54}{9} & \TthreeShade{TthreeGray!20}{3.47} & \TthreeShade{TthreeGray!20}{3.41} & \TthreeShade{TthreeGray!20}{3.65} \\
            Kimi-K2.7-Code & \TthreeShade{TthreeGray!100}{20} & \TthreeShade{TthreeGray!100}{12} & \TthreeShade{TthreeGray!100}{13} & \TthreeShade{TthreeGray!100}{13} & \TthreeShade{TthreeGray!100}{13} & \TthreeShade{TthreeGray!41}{3.75} & \TthreeShade{TthreeGray!39}{3.60} & \TthreeShade{TthreeGray!26}{3.70} \\
            Kimi-K3 & \TthreeShade{TthreeGray!37}{9} & \TthreeShade{TthreeGray!60}{9} & \TthreeShade{TthreeGray!54}{9} & \TthreeShade{TthreeGray!54}{9} & \TthreeShade{TthreeGray!54}{9} & \TthreeShade{TthreeGray!100}{4.56} & \TthreeShade{TthreeGray!100}{4.22} & \TthreeShade{TthreeGray!100}{4.33} \\
            \midrule
            GPT-5.6-Luna & \TthreeShade{white}{46} & \TthreeShade{white}{22} & \TthreeShade{white}{25} & \TthreeShade{white}{33} & \TthreeShade{white}{40} & \TthreeShade{white}{3.57} & \TthreeShade{white}{3.74} & \TthreeShade{white}{3.93} \\
            GPT-5.6-Terra & \TthreeShade{white}{46} & \TthreeShade{TthreeBlue!30}{24} & \TthreeShade{TthreeBlue!28}{27} & \TthreeShade{TthreeOrange!25}{31} & \TthreeShade{white}{40} & \TthreeShade{TthreeBlue!19}{3.70} & \TthreeShade{TthreeOrange!10}{3.70} & \TthreeShade{white}{3.93} \\
            GPT-5.6-Sol & \TthreeShade{TthreeBlue!21}{48} & \TthreeShade{TthreeBlue!98}{43} & \TthreeShade{TthreeBlue!85}{43} & \TthreeShade{TthreeBlue!60}{45} & \TthreeShade{TthreeBlue!45}{48} & \TthreeShade{TthreeBlue!43}{4.23} & \TthreeShade{TthreeBlue!31}{4.10} & \TthreeShade{TthreeBlue!29}{4.27} \\
            \midrule
            GPT-6-Luna & \TthreeShade{TthreeBlue!21}{48} & \TthreeShade{TthreeBlue!100}{44} & \TthreeShade{TthreeBlue!87}{44} & \TthreeShade{TthreeBlue!67}{48} & \TthreeShade{TthreeBlue!45}{48} & \TthreeShade{TthreeBlue!39}{4.10} & \TthreeShade{TthreeBlue!30}{4.08} & \TthreeShade{TthreeBlue!24}{4.15} \\
            GPT-6-Sol & \TthreeShade{TthreeBlue!21}{48} & \TthreeShade{TthreeBlue!100}{46} & \TthreeShade{TthreeBlue!92}{46} & \TthreeShade{TthreeBlue!67}{48} & \TthreeShade{TthreeBlue!45}{48} & \TthreeShade{TthreeBlue!44}{4.25} & \TthreeShade{TthreeBlue!33}{4.15} & \TthreeShade{TthreeBlue!29}{4.25} \\
            GPT-6-Astra & \TthreeShade{TthreeBlue!29}{\textbf{50}} & \TthreeShade{TthreeBlue!100}{\textbf{48}} & \TthreeShade{TthreeBlue!96}{\textbf{48}} & \TthreeShade{TthreeBlue!72}{\textbf{50}} & \TthreeShade{TthreeBlue!50}{\textbf{50}} & \TthreeShade{TthreeBlue!47}{\textbf{4.36}} & \TthreeShade{TthreeBlue!36}{\textbf{4.22}} & \TthreeShade{TthreeBlue!39}{\textbf{4.54}} \\
            \bottomrule
        \end{tabular}%
    }
\end{table*}

\noindent \textbf{Quantitative results on MaLiang-IBench.}
Tables~\ref{tab:llm_image_benchmark} and~\ref{tab:llm_image_delivery_quality} distinguish generation success from satisfaction of visual requirements. GPT-5.6-Luna and GPT-5.6-Terra each successfully generate 46 of 50 images. However, the number satisfying all three quality criteria falls to 22 for Luna and 24 for Terra.
\textbf{GPT-6-Astra achieves a 100\% generation success rate with 48 images satisfying all criteria.} GPT-6-Sol and GPT-6-Luna satisfy all criteria on 46 and 44 tasks. All successfully generated images from the three GPT-6 models meet the aesthetics and composition thresholds. Their remaining quality failures concern prompt adherence. This pattern shows that successful generation alone is insufficient to assess executable visual programs.

Besides, mean quality scores must be interpreted alongside the number of successful tasks. DeepSeek and Kimi generate only 6-20 images, so their means describe a limited subset of MaLiang-IBench. Kimi-K3's alignment score of 4.56 is based on just nine images and does not establish superior performance over the full evaluation set. Astra obtains the highest mean score in each quality dimension among the GPT models. The cost results reveal a trade-off between generation time and the number of images satisfying all criteria. GPT-5.6-Sol produces 43 qualifying images at 3.28 minutes per image. Astra produces 48 at 3.70 minutes per image. These time estimates include failures and retries and thus account for the cost of obtaining images that meet the quality thresholds.

\begin{figure}[t!]
    \centering
    \includegraphics[width=1.0\linewidth]{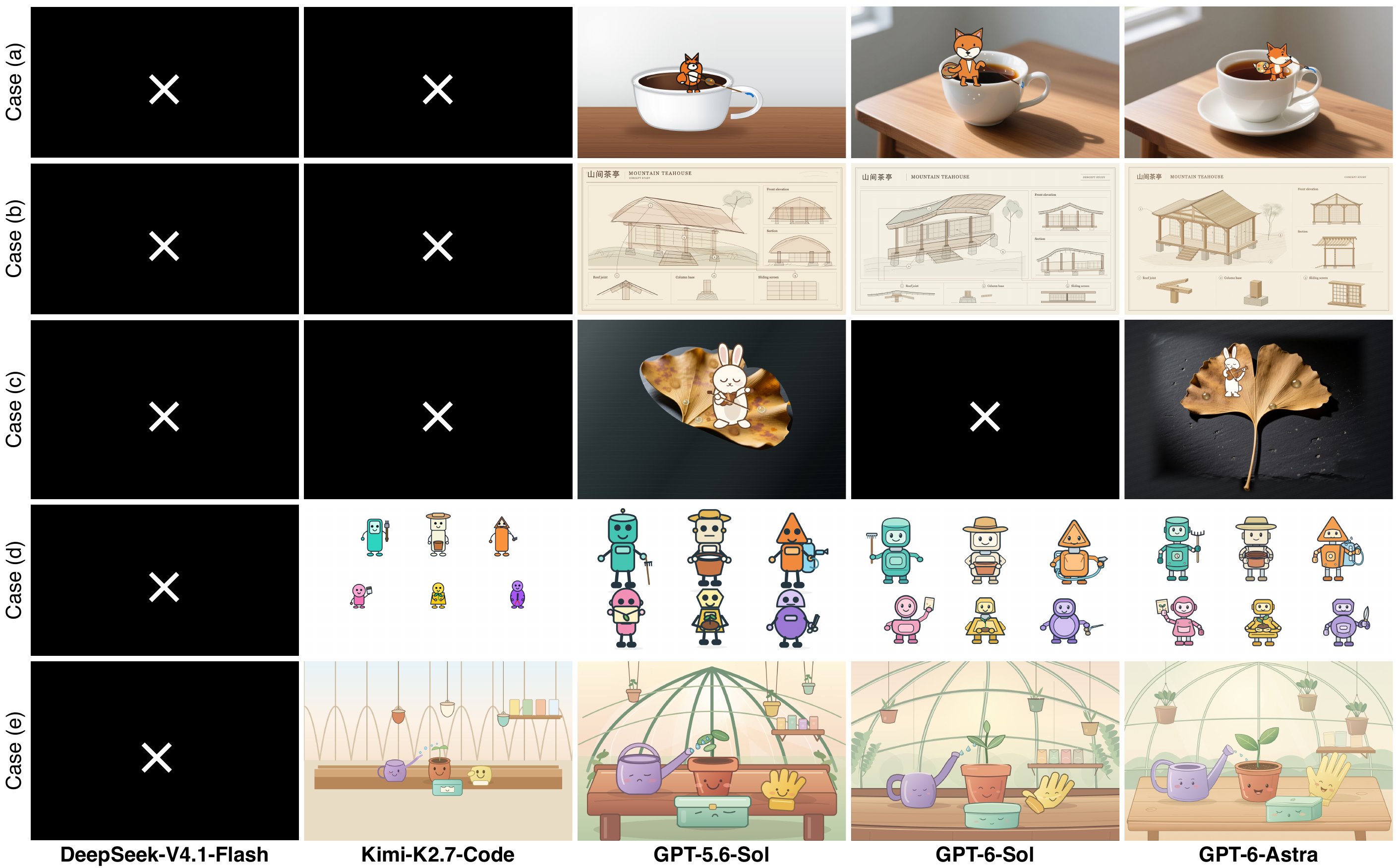}
    \caption{\textbf{Qualitative comparison on MaLiang-IBench.} Rows show a fox on a teacup, an architectural concept board, a rabbit on a ginkgo leaf, a robot ensemble, and a greenhouse illustration. Columns compare five models under the same prompt. A white $\times$ on black denotes an unsuccessful run. Full prompts are provided in Appendix~\ref{sec:qualitative_image_prompts}.}
    \label{fig:t2i_visual_result}
\end{figure}

\noindent \textbf{Qualitative results on MaLiang-IBench.}
Fig.~\ref{fig:t2i_visual_result} illustrates differences in visual realization across the evaluated models. The greenhouse (last row) and robot ensemble (fourth row) examples reveal differences in subject scale, scene detail, and the depiction of individual objects across models.
The teacup (first row) and ginkgo-leaf examples (third row) further probe the integration of flat cartoon characters with realistic materials.
Astra's results show more pronounced surface detail and lighting cues in these examples, while the architectural board combines a main view with supporting elevations and construction details.
These selected examples complement the aggregate scores by showing how the same prompt can lead to different compositions and levels of visual detail.

\begin{table*}[t!]
    \centering
    \definecolor{TfourBlue}{RGB}{183,203,242}
    \definecolor{TfourOrange}{RGB}{248,215,199}    \def\TfourShade#1#2{\begingroup\setlength{\fboxsep}{2pt}\colorbox{#1}{\makebox[3.2em]{#2}}\endgroup}
    \caption{\textbf{Generation success and computational cost on MaLiang-VBench.} ``Orange/blue'' denotes ``higher/lower'' values relative to DeepSeek-V4.1-Flash, except ``Time/Success'', which uses Kimi-K2.6. ``--'' indicates an undefined ratio with no successful generations.}
    \label{tab:llm_video_operational}
    \resizebox{\textwidth}{!}{%
        \begin{tabular}{lccccccccc}
            \toprule
            & \multicolumn{3}{c}{\textbf{Task Completion}}
            & \multicolumn{3}{c}{\textbf{Generation Time}}
            & \multicolumn{3}{c}{\textbf{Model Usage}} \\
            \cmidrule(lr){2-4}\cmidrule(lr){5-7}\cmidrule(lr){8-10}
            \textbf{Model} & Success $\uparrow$ & Failures $\downarrow$ & Token Limit $\downarrow$
            & Total Time & Time/Success $\downarrow$
            & Time/Task & Calls & Input Tokens & Output Tokens \\
            & (\%) & (cases) & (cases) & (task-h) & (min/video)
            & (min/case) & (calls/case) & (k/case) & (k/case) \\
            \midrule
            DeepSeek-V4.1-Flash (64K) & \TfourShade{white}{0.0} & \TfourShade{white}{13} & \TfourShade{white}{5} & \TfourShade{white}{1.47} & -- & \TfourShade{white}{6.77} & \TfourShade{white}{6.38} & \TfourShade{white}{560.5} & \TfourShade{white}{84.2} \\
            Kimi-K2.6 & \TfourShade{TfourOrange!48}{23.1} & \TfourShade{TfourBlue!48}{10} & \TfourShade{TfourBlue!89}{1} & \TfourShade{TfourOrange!100}{3.35} & \TfourShade{white}{66.95} & \TfourShade{TfourOrange!100}{15.45} & \TfourShade{TfourOrange!100}{16.15} & \TfourShade{TfourBlue!14}{548.8} & \TfourShade{TfourBlue!82}{26.9} \\
            GPT-5.6-Sol & \TfourShade{TfourOrange!73}{53.8} & \TfourShade{TfourBlue!73}{6} & \TfourShade{TfourBlue!63}{3} & \TfourShade{TfourBlue!52}{1.08} & \TfourShade{TfourBlue!93}{9.27} & \TfourShade{TfourBlue!51}{4.99} & \TfourShade{TfourOrange!100}{12.85} & \TfourShade{TfourOrange!16}{574.2} & \TfourShade{TfourBlue!91}{13.8} \\
            GPT-6-Astra & \TfourShade{TfourOrange!100}{100.0} & \TfourShade{TfourBlue!100}{0} & \TfourShade{TfourBlue!100}{0} & \TfourShade{TfourOrange!69}{2.17} & \TfourShade{TfourBlue!92}{10.00} & \TfourShade{TfourOrange!69}{10.00} & \TfourShade{TfourOrange!88}{11.31} & \TfourShade{TfourBlue!42}{463.3} & \TfourShade{TfourBlue!91}{14.4} \\
            \bottomrule
        \end{tabular}%
    }
\end{table*}

\noindent \textbf{Quantitative results on MaLiang-VBench.}
Tables~\ref{tab:llm_video_operational} and~\ref{tab:llm_video_qualified_delivery} show a similar gap between generation success and quality on MaLiang-VBench. \textbf{Astra satisfies all four quality thresholds on 10 of its 13 successful generations.} GPT-5.6-Sol satisfies all thresholds on five of its seven successful generations. Kimi-K2.6 completes three tasks but none of its videos meets all criteria. DeepSeek-V4.1-Flash completes no tasks. Token-budget exhaustion accounts for five of DeepSeek's 13 failures and one of Kimi's 10 failures. It also accounts for three of the six failures from GPT-5.6-Sol. Token limits therefore explain only part of the observed failures.

The quality results indicate that appearance alone does not capture temporal requirement satisfaction. All 13 Astra videos meet the aesthetics threshold. Only 10 meet the motion-coherence threshold, making motion the most restrictive criterion for Astra. All seven successful GPT-5.6-Sol videos meet the alignment and aesthetics thresholds. Requiring composition and motion coherence reduces the joint count to five. Astra achieves full task completion at 10.00 minutes per successful video. GPT-5.6-Sol requires 9.27 minutes per successful video but completes fewer tasks. ``Time/Success'' for videos uses the number of successful generations as its denominator. ``Time/Qualified'' for images instead uses the number of samples satisfying all quality thresholds.

\begin{table*}[t]
    \centering
    \definecolor{TfiveBlue}{RGB}{183,203,242}
    \definecolor{TfiveOrange}{RGB}{248,215,199}
    \definecolor{TfiveGray}{RGB}{205,205,205}
    \def\TfiveShade#1#2{\begingroup\setlength{\fboxsep}{2pt}\colorbox{#1}{\makebox[3.2em]{#2}}\endgroup}
    \caption{\textbf{Visual quality on MaLiang-VBench.} Quality counts include samples scoring at least 4 on a five-point scale. Gray denotes reference results. Bold marks the best GPT results excluding evaluation coverage. ``Blue/orange'' denotes ``higher/lower'' counts than GPT-5.6-Sol.}
    \label{tab:llm_video_qualified_delivery}
    \resizebox{\textwidth}{!}{%
        \begin{tabular}{lccccccc}
            \toprule
            & \multicolumn{2}{c}{\textbf{Generation Outcomes}}
            & \multicolumn{4}{c}{\textbf{Quality Threshold Counts}}
            & \multicolumn{1}{c}{\textbf{Evaluation Coverage}} \\
            \cmidrule(lr){2-3}\cmidrule(lr){4-7}\cmidrule(lr){8-8}
            \textbf{Model} & Successful $\uparrow$ & All Criteria $\uparrow$
            & Alignment $\uparrow$ & Aesthetics $\uparrow$
            & Composition $\uparrow$ & Motion $\uparrow$ & Evaluated \\
            & (cases) & (cases) & (cases) & (cases) & (cases) & (cases) & (cases) \\
            \midrule
            DeepSeek-V4.1-Flash (64K) & \TfiveShade{TfiveGray!20}{0} & \TfiveShade{TfiveGray!20}{0} & \TfiveShade{TfiveGray!20}{0} & \TfiveShade{TfiveGray!20}{0} & \TfiveShade{TfiveGray!20}{0} & \TfiveShade{TfiveGray!20}{0} & \TfiveShade{TfiveGray!20}{0} \\
            Kimi-K2.6 & \TfiveShade{TfiveGray!100}{3} & \TfiveShade{TfiveGray!20}{0} & \TfiveShade{TfiveGray!20}{0} & \TfiveShade{TfiveGray!100}{1} & \TfiveShade{TfiveGray!100}{1} & \TfiveShade{TfiveGray!20}{0} & \TfiveShade{TfiveGray!100}{3} \\
            \midrule
            GPT-5.6-Sol & \TfiveShade{white}{7} & \TfiveShade{white}{5} & \TfiveShade{white}{7} & \TfiveShade{white}{7} & \TfiveShade{white}{6} & \TfiveShade{white}{6} & \TfiveShade{white}{7} \\
            GPT-6-Astra & \TfiveShade{TfiveBlue!93}{\textbf{13}} & \TfiveShade{TfiveBlue!100}{\textbf{10}} & \TfiveShade{TfiveBlue!85}{\textbf{12}} & \TfiveShade{TfiveBlue!93}{\textbf{13}} & \TfiveShade{TfiveBlue!100}{\textbf{12}} & \TfiveShade{TfiveBlue!82}{\textbf{10}} & \TfiveShade{TfiveBlue!93}{13} \\
            \bottomrule
        \end{tabular}%
    }
\end{table*}

\begin{figure}[t!]
    \centering
    \includegraphics[width=1.0\linewidth]{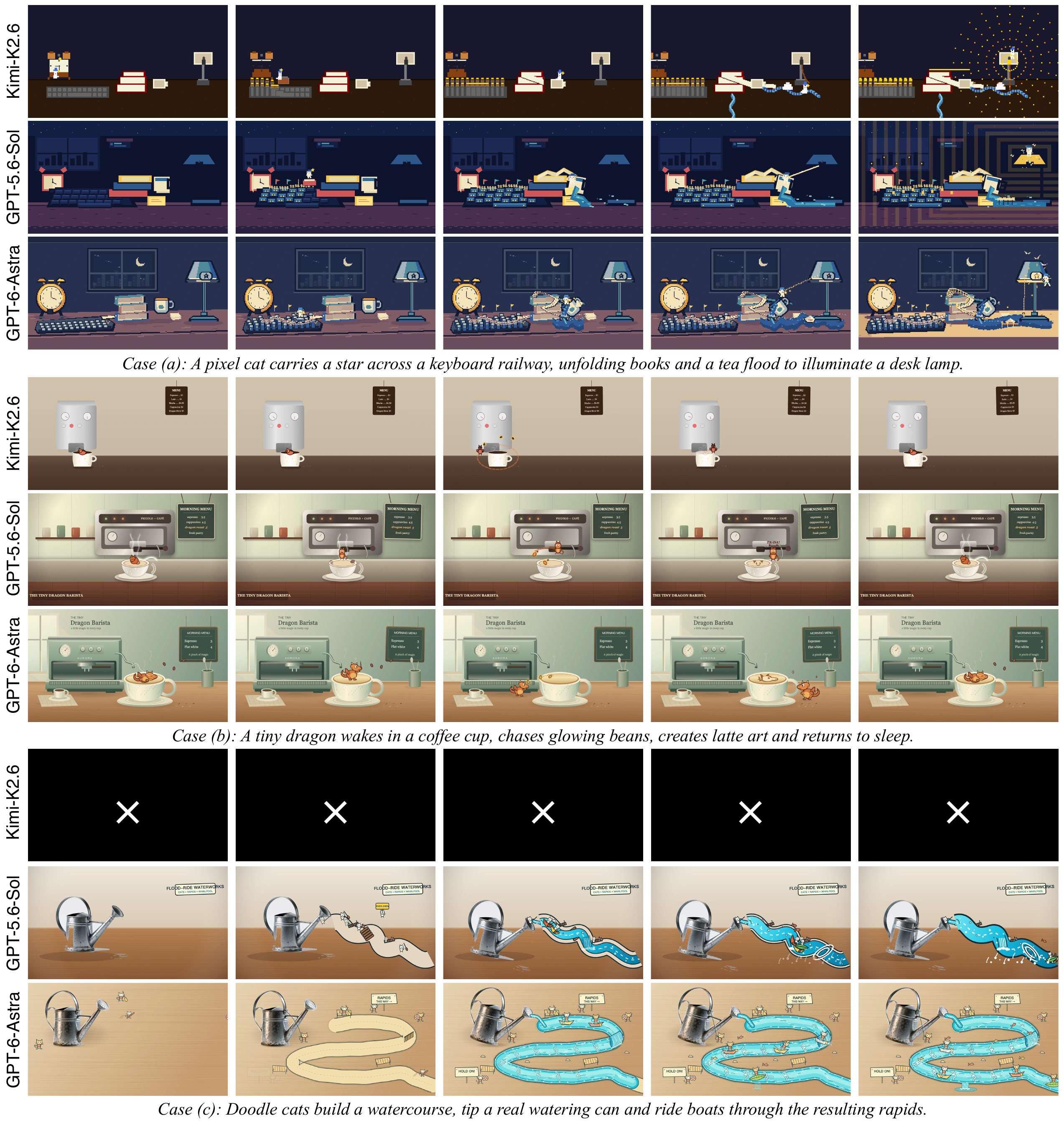}
    \caption{\textbf{Qualitative comparison on MaLiang-VBench.} Groups show (a) Star Delivery, (b) The Tiny Dragon Barista, and (c) Watering-Can Rafting. Each model row contains five uniformly sampled frames in temporal order from left to right. A white $\times$ on black denotes an unsuccessful run. Full prompts are provided in Appendix~\ref{sec:qualitative_video_prompts}.}
    \label{fig:t2v_visual_result}
\end{figure}

\noindent \textbf{Qualitative results on MaLiang-VBench.}
Fig.~\ref{fig:t2v_visual_result} compares the temporal progression of three selected video tasks.
In case (a), Astra retains a detailed pixel-art setting as the kitten traverses the keyboard and books toward the lamp, whereas Kimi depicts a simpler scene and a reduced action sequence.
Case (b) similarly contrasts Astra's more detailed setting and visible latte-art outcome with Kimi's sparse composition.
In case (c), both successful models depict a progression from constructing the channel to filling it with water and moving boats downstream, using different spatial arrangements.
Together, these examples highlight differences in both scene detail and the extent to which models realize the sequence of actions specified in the prompt.

\subsection{Discussion}

\begin{figure}[ht]
    \centering
    \includegraphics[width=1.0\linewidth]{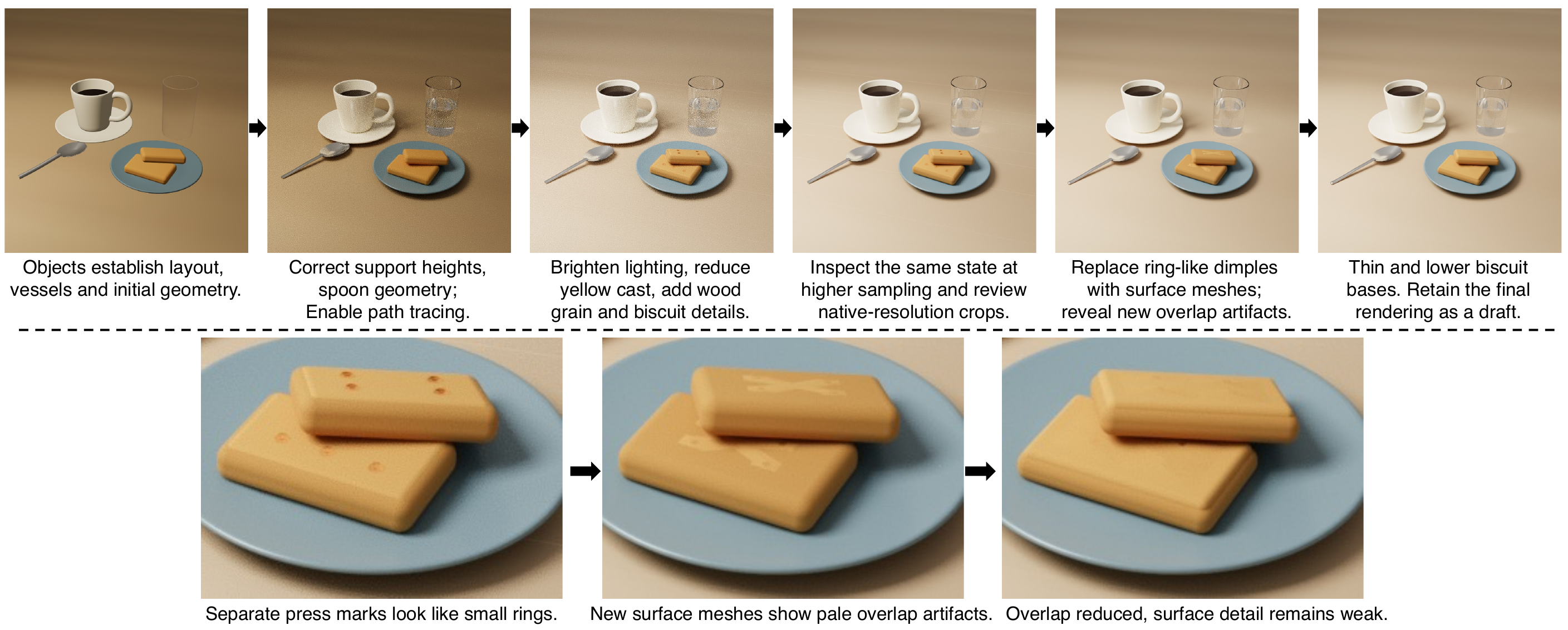}
    \caption{\textbf{Iterative refinement with a path-tracing backend.} Top: GPT-6-Astra revises a procedural breakfast scene through geometry, lighting, material, and sampling adjustments. Bottom: detail crops show ``biscuit'' surface artifacts introduced during refinement and reduced by geometry corrections.}
    \label{fig:pathtrace_refinement}
\end{figure}

\begin{figure}[ht]
    \centering
    \includegraphics[width=1.0\linewidth]{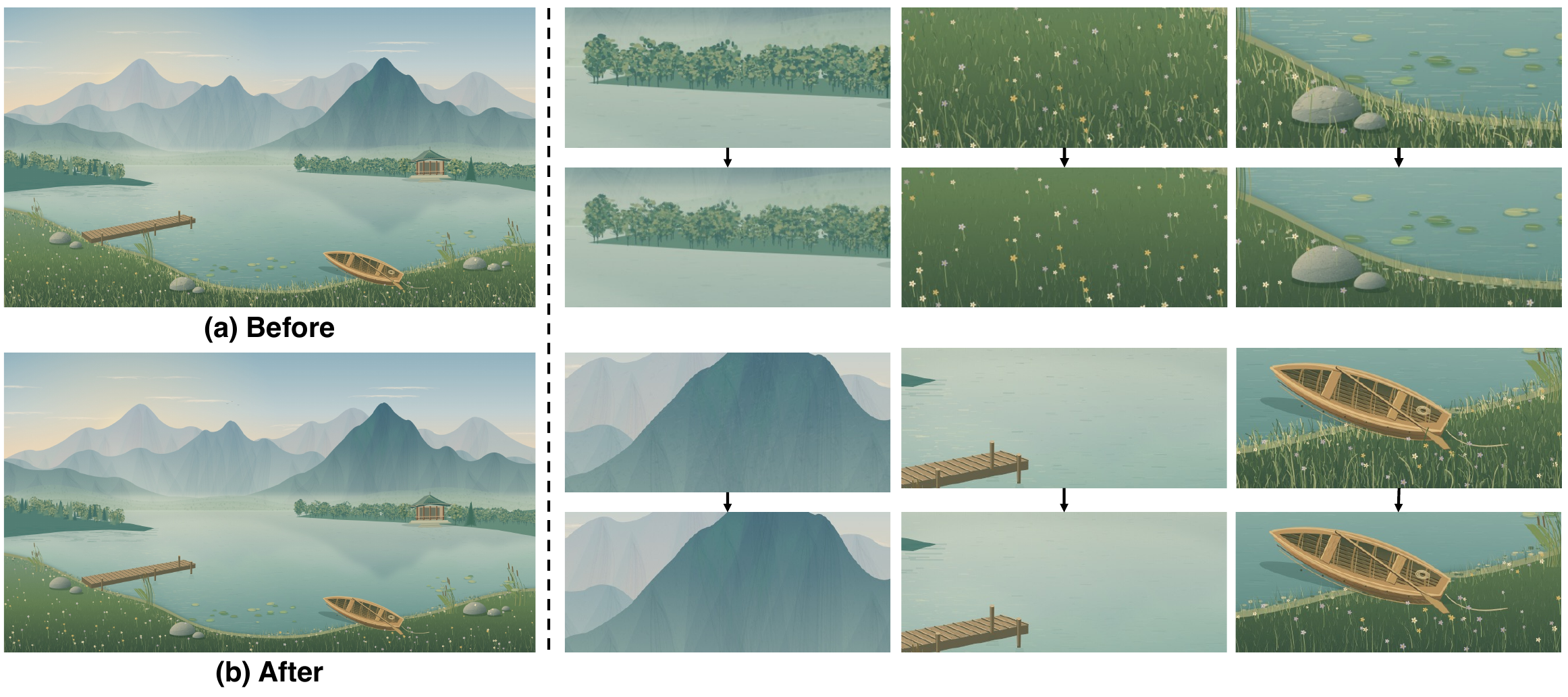}
    \caption{\textbf{Limits of prompted brushwork refinement.} Left: a Canvas illustration before and after GPT-6-Astra edits its drawing program to produce finer brushwork within REV. Right: paired detail crops show each region before (top) and after (bottom) editing. Finer marks do not yield photorealistic appearances: grass coverage and distinctive surface texture are instead reduced in several regions.}
    \label{fig:photorealistic_refinement}
\end{figure}

\paragraph{Can the harness generate photorealistic content?}
We explore two complementary directions toward more realistic visual content: extending the rendering backend and refining the brushwork of an existing illustration:
\textit{1) More expressive rendering backends.}
MaLiang-Harness can incorporate renderers with richer material and light-transport capabilities through its shared interface, while retaining the high-level planning and generation control described in Sec.~\ref{sec:unified_creation_interface}.
Fig.~\ref{fig:pathtrace_refinement} illustrates this approach with a path-tracing backend that supports structured scene geometry, physical materials, and area lighting. GPT-6-Astra constructs a breakfast still life without generated image assets, using a raster preview to establish composition before inspecting path-traced renderings. The resulting scene exhibits material-dependent reflections, glass transmission, and contact shadows. Further edits adjust object support, lighting, and surface geometry, while increasing the final sample count from 512 to 1024 reduces visible noise.
These refinements also reveal how local edits can introduce new visual defects. In the biscuit crops, a geometry change creates overlap artifacts that a subsequent revision reduces. PEG preserves the scene revisions, while TGP and REV make the changes and their visual consequences available for comparison and correction.
This example demonstrates the integration of a richer renderer within the same refinement process, while controlled comparisons are still needed to establish its contribution to perceptual realism.

\noindent \textit{2) Brushwork refinement inspired by hyperrealism.}
A complementary approach draws on the use of fine brushwork in hyperrealist painting. We examine whether an explicit refinement prompt, applied through REV, can move an existing stylized illustration toward a more realistic appearance without changing its Canvas backend. Fig.~\ref{fig:photorealistic_refinement} compares the inherited landscape with the preview produced after GPT-6-Astra modifies its drawing routines. The overall composition is preserved, while foliage, grass, stone, water, and timber receive smaller marks or thinner contours.
\textbf{However, the image remains stylized, and several regions lose existing detail}: grass coverage becomes less visible, while distinctive stone markings and mountain textures are attenuated. Finer strokes alone do not recover the coordinated shape, shading, and texture required for realistic depiction.
It highlights the need to preserve meaningful visual detail when translating refinement instructions into code edits.

\begin{figure*}[ht]
    \centering
    \begin{minipage}[t]{0.48\textwidth}
        \vspace{0pt}\centering
        \includegraphics[pagebox=cropbox,height=0.70\linewidth]{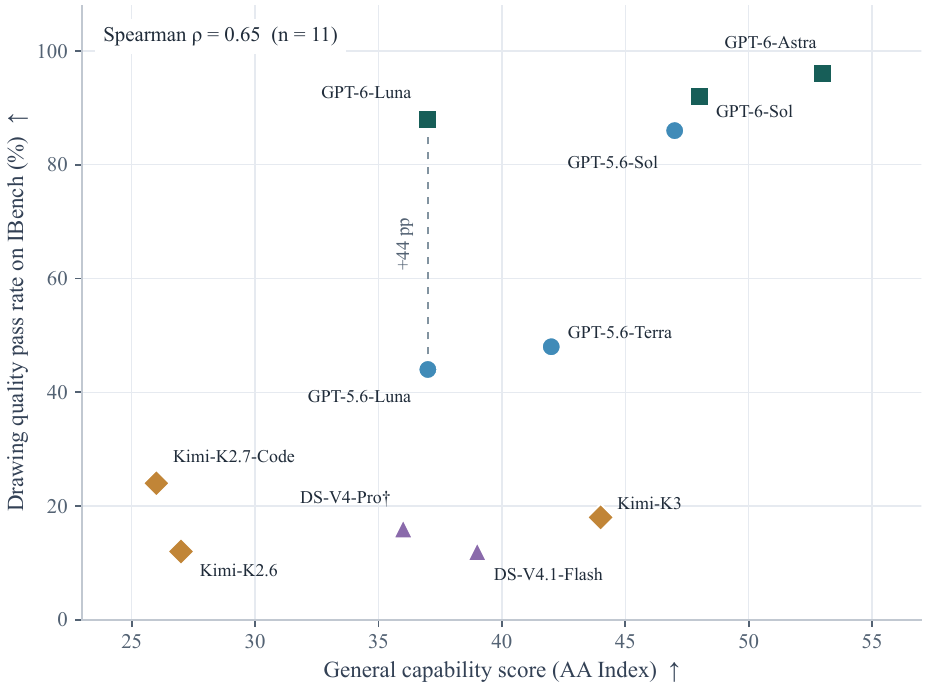}
        \par\smallskip
        \textbf{(a) General capability and drawing quality}
    \end{minipage}\hfill
    \begin{minipage}[t]{0.50\textwidth}
        \vspace{0pt}\centering
        \includegraphics[pagebox=cropbox,height=0.672\linewidth]{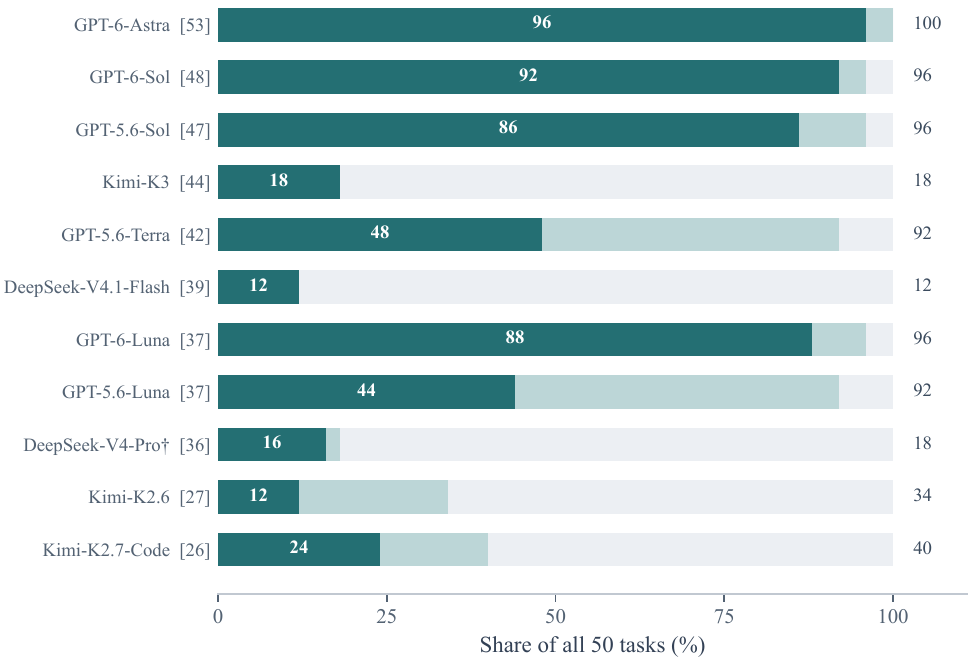}
        \par\smallskip
        \textbf{(b) Generation outcomes by model}
    \end{minipage}
    \caption{\textbf{General capability and visual program generation on MaLiang-IBench.} (a) Public AA Intelligence Index scores versus the percentage of all 50 tasks meeting every visual quality threshold. The dashed line highlights the \textbf{44-percentage-point gap} between the two Luna models. (b) Models ordered by AA score (brackets). Dark teal, light teal, and gray denote all criteria met, successful generation with unmet criteria, and failure. White numbers give quality pass rates and rightmost numbers give generation success rates. $^{\dagger}$The public V4-Pro score uses checkpoint 0813.}
    \label{fig:general_capability_visual_generation}
\end{figure*}

\paragraph{Does general MLLM capability predict visual program generation performance?}
We examine how general model capability relates to visual program generation by comparing public Artificial Analysis (AA) Intelligence Index scores~\citep{artificialanalysis2026intelligence} with MaLiang-IBench outcomes. \textit{We use the displayed integer scores from version 4.3.2, accessed on September 27, 2026, for the reasoning configurations listed on the model pages.} The index aggregates evaluations of knowledge, reasoning, coding, and agentic tasks. Drawing quality pass rate is the proportion of all 50 tasks that satisfy the alignment, aesthetics, and composition thresholds, including unsuccessful generations in the denominator. Across the 11 models, the two measures exhibit a positive rank correlation (Spearman $\rho=0.65$ in Fig.~\ref{fig:general_capability_visual_generation}a).
\textbf{However, general capability scores do not fully predict visual performance.}
GPT-5.6-Luna and GPT-6-Luna share a displayed index score of 37, yet their drawing quality pass rates are 44\% and 88\%. Kimi-K3 scores 44 on the index but satisfies all visual criteria on only 18\% of tasks. Fig.~\ref{fig:general_capability_visual_generation}b further separates completion from visual quality: GPT-5.6-Luna and GPT-5.6-Terra each complete 92\% of tasks, while only 44\% and 48\% meet all quality thresholds. These observations suggest that general benchmark performance captures only part of the demands of visual program generation, which also requires translating spatial and appearance constraints into executable representations and using rendered feedback to guide revision.

\begin{figure*}[t]
    \centering
    \begin{minipage}[t]{0.52\textwidth}
        \vspace{0pt}
        \centering
        \textbf{(a) Rendered ``research-figure'' draft}\par\smallskip
        \includegraphics[width=\linewidth]{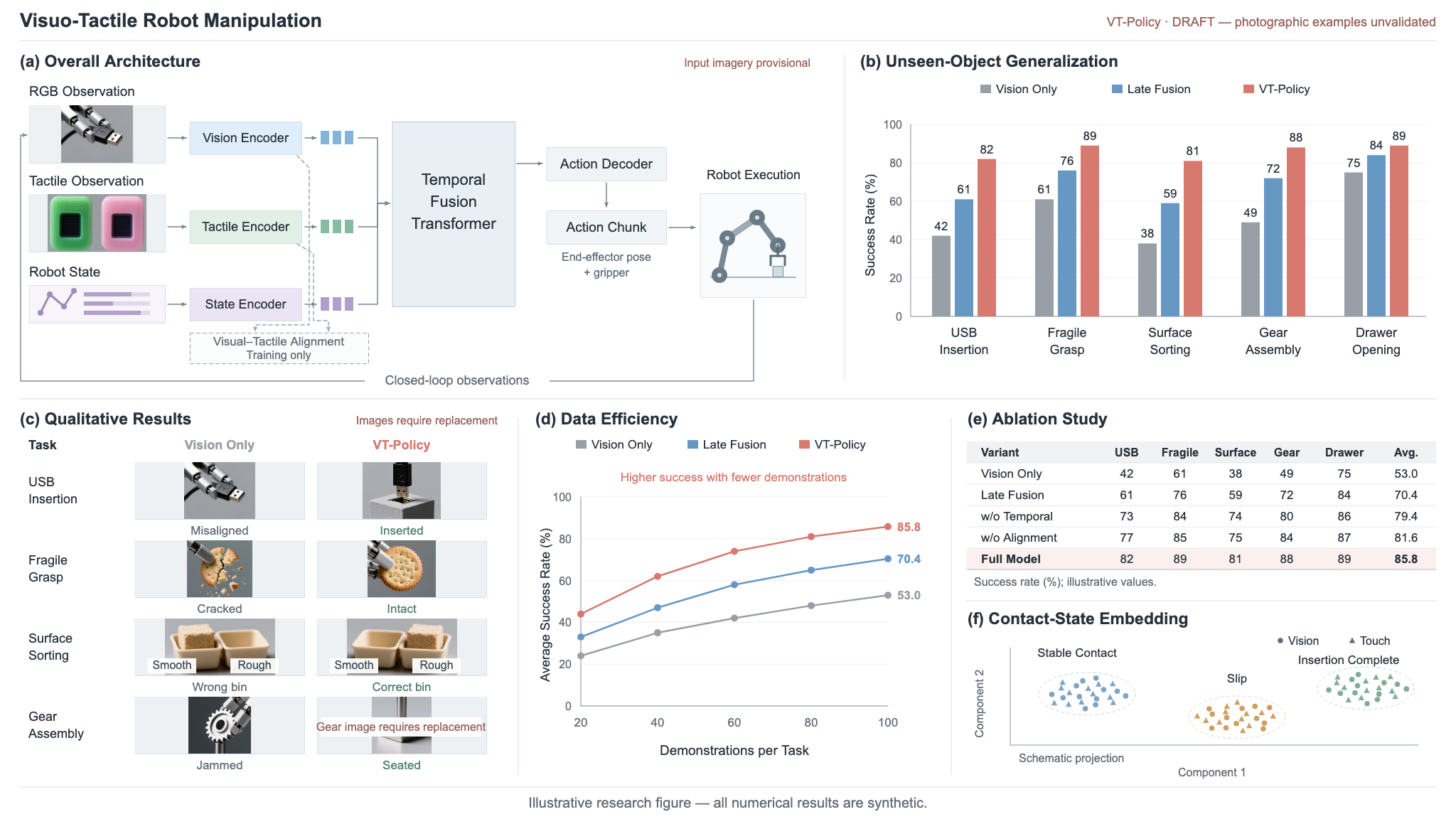}
    \end{minipage}\hfill
    \begin{minipage}[t]{0.46\textwidth}
        \vspace{0pt}
        \centering
        \textbf{(b) Construction trace in pseudocode}\par\smallskip
        \raggedright
        \scriptsize
        \begin{algorithmic}[1]
            \Require Figure brief $B$, synthetic data $D$
            \Statex \textbf{Programmatic construction}
            \State $S \gets \Call{InitCanvas}{2048,1152}$
            \State Draw architecture and arrows in $S$. \Comment{a}
            \State Draw bars from $D$; reserve photo cells. \Comment{b,c}
            \State Draw learning curves and ablation table. \Comment{d,e}
            \State Draw schematic contact-state clusters. \Comment{f}
            \Statex \textbf{Asset generation and selection}
            \State $L\gets[2\!\times\!5,\,2\!\times\!5,\,3\!\times\!3,\,2\!\times\!4]$
            \For{$k=1,\ldots,4$}
                \State $A_k\gets\Call{GenerateAssets}{B,L_k}$
                \State Inspect $A_k$ against photo requirements.
                \State Record provenance; revise the next prompt.
            \EndFor
            \State $A^*\gets A_3$ \Comment{choice in this run}
            \Statex \textbf{Composition and verification}
            \State Crop and place $A^*$ in the reserved cells.
            \State Overlay labels using code.
            \State $I\gets\Call{Render}{S}$
            \State Inspect the full image and detail crops.
            \State Add notices for unresolved photo requirements.
            \State $I\gets\Call{Render}{S}$
            \State $Q\gets\Call{Review}{I,B}$
            \State Export $I$ as PNG; retain draft status under $Q$.
        \end{algorithmic}
    \end{minipage}
    \caption{\textbf{Tracing a ``research figure case'' to its construction process.} Left: a generated six-panel ``research-figure'' poster. Right: case-specific pseudocode abstracted from the execution trace (comments a-f refer to panels in the rendered image). The recorded program, asset-generation steps, and reviews make it possible to trace how individual elements were constructed and assessed.}
    \label{fig:interpretable_construction}
\end{figure*}

\paragraph{What can the construction process reveal beyond the final image?}
MaLiang-Harness makes the construction of a visual result inspectable through its executable program, asset provenance, and revision history. Fig.~\ref{fig:interpretable_construction} pairs a ``six-panel research illustration'' of an academic paper with pseudocode summarizing its recorded construction process.
GPT-6-Astra uses code to specify the layout, architecture diagram, labels, and plots from user-supplied synthetic data, while generating photographic-style assets for the manipulation examples. The trace records four asset-generation attempts with different contact-sheet layouts.
The final composition uses crops from the third attempt, with code controlling their placement, scaling, and labels. The construction record distinguishes elements drawn from explicit data and instructions from those assembled using generated imagery.
\textbf{This distinction also helps interpret the final review}: the numerical plots and layout pass, whereas the photographic examples remain unsatisfactory, leaving the output only as a draft. By linking construction steps to revision-specific assessments, the harness reveals how individual elements were produced and which requirements remain unresolved beyond what the final image alone shows.

\paragraph{When does refinement stall, and when should it stop?}
The harness faces three potential failure modes: \textbf{reasoning stagnation}, where planning does not advance the artwork; \textbf{semantic livelock}, where repeated edits leave requirements unresolved; and \textbf{infinite agentic loops}, where planning and tool use repeat without termination.
Refinement stalls when continued activity produces no meaningful visual improvement. PEG preserves the working state, while TGP and REV support inspection, comparison, and recovery, but these mechanisms do not guarantee effective corrections. In Fig.~\ref{fig:photorealistic_refinement}, further inspection yields no corrective edit before the token budget is exhausted.
The harness therefore separates termination from completion: model-call, tool-call, time, and token limits bound execution, while successful finalization is required to mark a task completed.

\section{Conclusion}
\label{sec:applications}

We presented MaLiang-Harness, a unified framework for programmable image and video generation motivated by the Program-to-Visual (P2V) gap between program-level correctness and visual requirement satisfaction. By maintaining executable state, recording construction histories, and linking verification to the current revision, the framework organizes MLLM-driven visual creation into a persistent process of construction, inspection, and refinement. Evaluation of 11 MLLMs on MaLiang-IBench and four on MaLiang-VBench reveals substantial differences in generation success, visual quality, and computational cost. Models with similar general-capability scores can produce markedly different visual outcomes, motivating direct evaluation of visual program generation. Photorealism and stalled refinement remain challenges. Future work will explore richer rendering backends and progress-aware refinement strategies to improve visual fidelity and generation efficiency.

\newpage
\bibliographystyle{plainnat}
\bibliography{neurips_2026}

\section{Appendix}
\label{sec:appendix}
This appendix provides an expanded construction example, evaluation and aggregation details, the computation of the comparative plots, and the prompts used for the qualitative examples.

\subsection{Procedural Construction Example}
\label{sec:procedural_construction_details}

Algorithm~\ref{alg:botanical-portrait} expands the botanical portrait example in Fig.~\ref{fig:peg-botanical}. It summarizes the drawing operations within a fixed PEG revision, rather than the full planning and refinement loop. The program builds the image in four layers: the background, circular foliage, the portrait, and foreground flowers. Separate seeded generators control variation in foliage, portrait texture, and flowers. The named routines abstract the underlying Canvas operations; no image assets are used. As in Algorithm~\ref{alg:botanical}, $k$ denotes the flower-layer index within this algorithm, rather than a PEG revision.

\begin{algorithm}[t]
\caption{Procedural generation of the botanical portrait}
\label{alg:botanical-portrait}
\begin{algorithmic}[1]
\Require Canvas size $W=1024$, $H=576$; fixed seed $s$
\Ensure Rendered RGB image $I$

\State $C \gets \Call{NewCanvas}{W,H}$
\State $R_f,R_w,R_p \gets
       \Call{SeededGenerators}{s,\text{foliage},\text{woman},\text{flowers}}$

\Comment{Layer 1: white background}
\State \Call{FillRectangle}{$C,(0,0,W,H),\text{white}$}

\Comment{Layer 2: circular foliage}
\For{$i=0$ to $25$}
    \State $\theta \gets 2\pi i/26+\Call{Jitter}{R_f}$
    \State $\mathbf{x},\mathbf{y} \gets
           \Call{BranchEndpoints}{(W/2,H/2),\theta,R_f}$
    \State \Call{StrokeBezierBranch}{$C,\mathbf{x},\mathbf{y}$}
    \For{each position along the branch}
        \State \Call{DrawGradientLeaves}{$C,R_f$}
        \State \Call{DrawLeafVeinsAndGrain}{$C,R_f$}
    \EndFor
\EndFor
\State \Call{DrawBerrySprigsAndAdditionalLeaves}{$C,R_f$}

\Comment{Layer 3: woman's portrait}
\State \Call{FillBezierShape}{$C,\text{hair silhouette},
                              \text{brown gradient}$}
\State \Call{FillBezierShape}{$C,\text{neck and shoulders},
                              \text{skin gradient}$}
\State \Call{FillBezierShape}{$C,\text{face},\text{skin gradient}$}
\State \Call{AddClippedShadingAndGrain}{$C,\text{face},R_w$}
\State \Call{DrawFacialPaths}{$C,\text{brows, closed eyes,
                                     lashes, nose, lips}$}
\State \Call{FillBezierShape}{$C,\text{front hair locks},
                              \text{brown gradient}$}
\For{$d\in\{\text{left},\text{right}\}$}
    \For{$j=0$ to $74$}
        \State \Call{StrokeClippedHairStrand}{$C,d,j,R_w$}
    \EndFor
\EndFor

\Comment{Layer 4: flowers and foreground details}
\For{each prescribed rose position $(x,y,r)$}
    \State \Call{DrawSepals}{$C,x,y,r$}
    \For{$k=0$ to $4$}
        \State $n\gets [8,7,6,5,4]_k$
        \For{$j=0$ to $n-1$}
            \State \Call{DrawRotatedBezierPetal}
                   {$C,x,y,r,k,j,R_p$}
            \State \Call{AddPetalGradientAndGrain}{$C,R_p$}
        \EndFor
    \EndFor
\EndFor
\For{each prescribed daisy or small-blossom position}
    \State \Call{DrawRadialPetalsAndCenter}{$C,R_p$}
\EndFor
\State \Call{DrawForegroundLeavesAndBerries}{$C,R_p$}

\State $I\gets\Call{Rasterize}{C}$
\State \Return $I$
\end{algorithmic}
\end{algorithm}

\subsection{Evaluation Protocol and Reporting Details}
\label{sec:eval_protocol_details}

\paragraph{Task outcomes and quality metrics.}
MaLiang-IBench contains 50 image tasks and MaLiang-VBench contains 13 video tasks. A task succeeds when its output is decodable and passes the harness's completion checks; all remaining tasks are counted as failures. \emph{Token Limit} identifies the subset of failures that exhaust the allocated token budget. These outcome categories are distinct from the subsequent quality assessment by GPT-6-Sol. Images are scored from 1 to 5 for prompt alignment, aesthetics, and composition; videos additionally receive a motion-coherence score. Each dimension's threshold count includes successful outputs scoring at least 4, and \emph{All Criteria} counts those meeting every threshold. Mean quality scores use successful outputs only. Success rates and quality pass rates use the full task set as their denominator, so failed tasks contribute to neither numerator.

\paragraph{Video review.}
A single reviewer evaluated all 23 successful videos across the four models using the original prompt and 12 sampled frames in temporal order. \emph{Evaluated} records the number of reviewed videos. DeepSeek's zero quality counts reflect the absence of successful outputs, not assigned scores of zero. The review was not formally blinded. Motion coherence measures the progression visible in the sampled frames and does not establish frame-by-frame fluidity or rule out brief intermediate faults.

\paragraph{Computational cost.}
Cumulative generation time sums the durations of recorded attempts, including failures and retries; it is not the wall-clock duration of a batch executed in parallel. \emph{Time/Qualified} divides cumulative time by the number of images meeting all three quality thresholds. \emph{Time/Success} divides cumulative time by the number of successful videos and is undefined when none succeeds. Input and output token counts come from available usage records and include repeated input context. Per-task statistics follow the aggregation rules below.

\paragraph{Image run configurations and aggregation.}
MaLiang-IBench combines multiple batches, retries, and historical runs rather than uniform first attempts. DeepSeek-V4-Pro is evaluated without visual feedback. GPT-6-Astra's totals cover all 50 prompts, combining 39 attempts on 38 new tasks with 12 historical runs. New runs use worker wall time; historical runs use elapsed generation time without worker startup. Astra's per-task statistics divide totals, including retries, by 50. Kimi's mean call count also uses 50 as the denominator, whereas its \emph{Time/Task} is the median over successful tasks and its token statistics divide recorded totals by the number of successful tasks. Thus, identically named cost columns do not always use identical aggregation rules, limiting direct efficiency comparisons.

\paragraph{Video run configurations and aggregation.}
The reported MaLiang-VBench results use one recorded attempt per model and task. Time, calls, and tokens per task are averaged over all 13 attempts. DeepSeek results use the completed evaluation with a 64K output-token limit; the incomplete 32K batch is excluded. Astra results use historical runs. Total time sums attempt wall times for new batches and elapsed generation times for historical runs. Differences in budgets, timing conventions, and run conditions mean that cost comparisons describe the evaluated configurations.

\subsection{Comparative Plots and Table Conventions}
\label{sec:comparative_plot_details}

\paragraph{Radar rankings.}
Fig.~\ref{fig:rank_radar} uses six metrics from Tables~\ref{tab:llm_image_benchmark} and~\ref{tab:llm_image_delivery_quality}: mean alignment, aesthetics, and composition scores; generation success rate; reported time per task; and the sum of input and output tokens per task. Each metric is ranked across all 11 models before grouping them into panels. Higher quality and success values rank better, whereas lower costs rank better. Ties in the reported values receive average ranks. A rank $r$ is mapped to radius $(12-r)/11$, placing rank 1 at the outer edge. Polygon area is not an aggregate performance score, and radial differences represent rank differences rather than differences in the underlying metrics. Legend counts indicate successful outputs. Small successful subsets can produce high quality means, while early failures can reduce costs; the aggregation differences described above also apply to this figure.

\paragraph{General-capability comparison.}
Fig.~\ref{fig:general_capability_visual_generation} uses the displayed integer Artificial Analysis Intelligence Index scores from version 4.3.2, accessed on September 27, 2026, for the reasoning configurations listed on the public model pages~\citep{artificialanalysis2026intelligence}. The image quality pass rate is the \emph{All Criteria} count divided by 50. Spearman correlation between this rate and the index is $\rho=0.65$ across 11 models. The public DeepSeek-V4-Pro score refers to checkpoint 0813, whose correspondence to our evaluated checkpoint is unverified; excluding it gives $\rho=0.63$. Public evaluation settings may differ from those used in our harness, so this analysis characterizes an association rather than isolating the effects of model capability.

\paragraph{Numerical color scales.}
In Tables~\ref{tab:llm_image_benchmark} and~\ref{tab:llm_video_operational}, orange and blue indicate values above and below DeepSeek-V4.1-Flash, respectively. Colors encode numerical differences, not a uniform preference across metrics. Relative-change intensity is $\sqrt{\min(|x/b-1|,1)}$, where $x$ is a reported value and $b$ is its baseline. Because DeepSeek has zero successful videos, video success-rate intensity instead uses $\sqrt{|x-b|/100}$, with $x$ and $b$ expressed as percentages. Video \emph{Time/Success} uses Kimi-K2.6 as its baseline.

\paragraph{Quality-table highlighting.}
In Tables~\ref{tab:llm_image_delivery_quality} and~\ref{tab:llm_video_qualified_delivery}, blue and orange indicate higher and lower GPT results relative to GPT-5.6-Luna for images and GPT-5.6-Sol for videos. Color intensity follows the relative-change rule above. DeepSeek and Kimi use a separate gray scale because their quality results cover fewer successful tasks; darker gray denotes larger values within each reference column. Bold identifies the best GPT values, excluding the descriptive \emph{Evaluated} column. Quality-threshold counts remain comparable as counts over the full task set, while means describe different successful subsets.

\subsection{Prompts for the Qualitative Image Comparisons}
\label{sec:qualitative_image_prompts}

The five prompts below follow the row order in Fig.~\ref{fig:t2i_visual_result}. Chinese prompts and display text are translated into English for presentation; evaluation used the original wording. Prompt content is presented separately from the evaluation protocol and does not imply that a generated output satisfies every requested property.

\medskip
\noindent\textbf{Fox Painter on a Teacup.}\par

A flat 2D cartoon fox painter stands on the rim of a realistic white porcelain teacup. Output a $1536\times864$ landscape PNG. Place the teacup slightly left of center on a wooden table, with natural highlights on its glaze, its handle on the right, and dark tea visible inside. The fox holds a paintbrush in one paw and a small palette in the other, painting a short blue arc on the handle. Both feet must rest on the rim nearest the viewer; the fox must not float or stand in the tea.

Render the fox with solid orange color blocks, clear black outlines, and a white muzzle, creating a distinct contrast with the realistic ceramic and wood grain. Use soft window light from the left, a light gray background, and negative space on the right. The brush tip must touch the handle, and the character must not be larger than the entire teacup. Do not add text, extra characters, or a frame around the entire scene.

\medskip
\noindent\textbf{Mountain Teahouse Concept.}\par

Create a $2048\times1152$ landscape architectural survey board for a fictional timber teahouse, titled ``Mountain Teahouse'' and ``MOUNTAIN TEAHOUSE'', with a small ``CONCEPT STUDY'' label.

Use a wide main panel on the left and a narrower column of two panels on the right. The left panel shows a three-quarter axonometric teahouse raised on stone piers on a hillside: one curved gable roof, an open front veranda, four evenly spaced front columns, sliding rear screens and a short stair descending toward the foreground. Add a small tree behind the building without hiding the roof.

The upper-right panel, ``Front elevation'', repeats the four columns, roof curve and central stair. The lower-right panel, ``Section'', cuts across the veranda and enclosed rear room, revealing roof rafters and stone supports. A bottom strip contains three enlarged details labeled ``Roof joint'', ``Column base'', ``Sliding screen''. Use leaders 1--3 to connect these features in the main view to their matching detail numbers. Do not add measurements or historical dates.

Use fine sepia pen lines, restrained watercolor washes, warm ivory paper, a thin double-line border and an orderly grid. This is a fictional conceptual documentation sheet, not a factual historical survey. Do not invent historical dates, engineering safety claims or paragraphs of unreadable text. Use only the exact short labels requested, with clear typography and thin leaders. No photographic rendering, bright neon colors or modern vehicles. Output one static PNG.

\medskip
\noindent\textbf{Rabbit Violinist on a Ginkgo Leaf.}\par

Draw a $1536\times1024$ landscape mixed-media illustration: a realistic golden-brown ginkgo leaf lies flat on damp, dark slate, with visible veins, curled edges, and exactly three separate transparent dewdrops. A white 2D cartoon rabbit stands on the broad upper portion of the leaf, playing a simplified small brown violin. Its left paw holds the violin neck, its right paw holds the bow, and the bow crosses the strings. The rabbit's eyes are gently closed.

Render the rabbit and violin with clear brown outlines, flat colors, and minimal shading; render the leaf, dewdrops, and slate with realistic materials and soft side lighting. The rabbit's body height is approximately one-third of the leaf's length, with a small contact shadow beneath its feet. Keep the background quiet, with no other leaves, text, musical notes, or characters. Output a static PNG.

\medskip
\noindent\textbf{Garden Robot Cast.}\par

Draw a $2048\times1152$ landscape robot cast illustration on a white background, containing exactly six original garden robots in two rows of three. In the back row, from left to right: a tall teal cylindrical robot holding a small rake; an off-white robot with a square head, wearing a wide-brimmed gardening hat and holding an empty flowerpot; and an orange robot with a triangular head, carrying a light blue water tank on its back and holding a spray hose. In the front row, from left to right: a short pink robot with a round head, raising a packet of seeds with no text; a yellow robot with a trapezoidal body, holding a seedling in both hands; and a purple robot with an oval body, holding a closed pair of pruning shears.

All robots have two eyes, two arms, and two legs, but distinct head and body silhouettes. Each tool belongs only to its assigned character. Make the back row taller and the front row shorter, slightly staggering the silhouettes without obscuring any character's face, hands, or feet. Use clean bold outlines, flat colors, minimal shading, and a children's animation style. Do not add a garden background, text, logos, extra robots, or scattered tools. Output PNG.

\medskip
\noindent\textbf{Greenhouse Potting Bench.}\par

Create a warm $1536\times1024$ landscape children's-book illustration of a greenhouse potting bench. The main cast consists of a smiling terracotta plant pot with a seedling, a shy lavender watering can, a cheerful yellow gardening glove, and a sleepy mint-green seed box. Each has exactly one face. Place the pot at the center, the watering can to its left with its spout pointing toward the pot, the glove resting to the right, and the closed seed box at the front edge of the bench.

In the background, curved greenhouse ribs frame a pale morning sky. Hang three faceless pots at different heights and show a short shelf with blank seed packets. A few drops of water are suspended between the watering-can spout and the seedling, suggesting a single frozen moment. Use soft peach, sage, butter yellow and cream, rounded silhouettes, fine outlines and gentle shading. Keep every character recognizable and do not draw human gardeners, text or extra anthropomorphic objects. Output PNG.

\paragraph{Additional example with explicit layout constraints.}
Beyond the qualitative examples in the main text, the following prompt illustrates how spatial constraints can be specified through object identifiers, normalized coordinates, dimensions, and color codes. These specifications are supplied as part of the text prompt and interpreted by the MLLM to construct an executable visual program.


Draw a $2048\times1152$, 16:9 landscape layout control diagram for a ``museum triptych exhibition board'' and output a static PNG. Draw only the colored rectangles and IDs specified below. Do not draw actual posters, people, objects, or final copy.

All rectangle coordinates are given as $[x,y,\mathrm{width},\mathrm{height}]$, with the horizontal and vertical axes mapped independently to a normalized range of 0--1000 using the actual canvas width and height. Do not mistakenly produce a square image, rearrange the layout, or automatically align the elements. Use a white background with no outlines, rounded corners, shadows, or gradients. Center each ID within its rectangle using a clear black sans-serif font. The list contains all rectangles; there are no unlisted parent containers or additional legends.

Color coding: coral red \texttt{\#FF6B6B} denotes titles; sky blue \texttt{\#62CBE8} denotes main illustration regions; magenta-purple \texttt{\#D86BDA} denotes photo regions; golden yellow \texttt{\#F4C95D} denotes analysis regions; mint green \texttt{\#72C7A0} denotes color palettes; and lavender \texttt{\#A999E8} denotes labels.

\begin{center}
\begin{tabular}{lll}
\toprule
ID & $[x,y,\mathrm{width},\mathrm{height}]$ & Color \\
\midrule
H1 & $[30,30,650,85]$ & Coral red \\
H2 & $[730,30,240,85]$ & Coral red \\
P1 & $[30,155,260,470]$ & Magenta-purple \\
T1 & $[30,645,260,45]$ & Lavender \\
M1 & $[320,155,390,330]$ & Sky blue \\
D1 & $[320,515,180,110]$ & Golden yellow \\
D2 & $[530,515,180,110]$ & Golden yellow \\
T2 & $[320,645,390,45]$ & Lavender \\
P2 & $[740,155,230,210]$ & Magenta-purple \\
P3 & $[740,395,230,230]$ & Magenta-purple \\
T3 & $[740,645,230,45]$ & Lavender \\
S1 & $[30,735,290,150]$ & Sky blue \\
S2 & $[355,735,290,150]$ & Sky blue \\
S3 & $[680,735,290,150]$ & Sky blue \\
F1 & $[30,925,940,40]$ & Lavender \\
\bottomrule
\end{tabular}
\end{center}

Spatial relationships: the main image in the middle column is wider than the left and right columns; the two photo regions in the right column have different heights; and the three bottom image regions have equal widths, with white gaps between them.

Check that there are exactly 15 rectangles, each ID appears exactly once, all coordinates and colors are accurate, and no rectangles overlap or extend beyond the canvas. Leave all unspecified gaps white. Do not add connecting lines, arrows, coordinate axes, borders, photographs, actual text content, or extra color blocks. Complete the task using only SVG or Canvas code.

\subsection{Prompts for the Qualitative Video Comparisons}
\label{sec:qualitative_video_prompts}

The following prompts correspond to Star Delivery, The Tiny Dragon Barista, and Watering-Can Rafting in Fig.~\ref{fig:t2v_visual_result}. Star Delivery is translated from Chinese for presentation; evaluation used the original Chinese wording. The other two prompts were originally written in English. The 12-frame quality review described above is distinct from the five uniformly sampled frames displayed for each model in the comparison figure.

\medskip
\noindent\textbf{Star Delivery: Lighting Up Desk City.}\par

A pixel kitten catches a star ejected from an alarm clock, rides a keyboard train across a desk, slides along book pages into rapids created by a teacup, and finally swings high into the air using a desk lamp's pull cord to place the star inside the unlit bulb.

\emph{Core action sequence:} the alarm clock ejects a star $\rightarrow$ a keyboard railway grows $\rightarrow$ a cart accelerates $\rightarrow$ the spacebar launches it $\rightarrow$ book pages unfold $\rightarrow$ tea floods the desk $\rightarrow$ a paper boat rides the rapids $\rightarrow$ a sweeping pull-cord swing $\rightarrow$ starlight illuminates the entire desk.

\emph{Ready-to-use prompt.} Create an imaginative animation entirely in pixel art, with clear action progression and a landscape 16:9 composition. The entire world must use a consistent, detailed 2D pixel-art style: characters, everyday objects, backgrounds, liquids, smoke, lighting, and particles must all consist of clearly recognizable square pixels. Use a uniform pixel size, a limited and harmonious palette, stepped contours, and a small amount of pixel dithering to convey depth. Do not use photorealistic assets, hand-drawn lines, or 3D voxels, and do not simply apply a mosaic filter to ordinary video.

The scene is a desk at night, viewed from the side at a slightly elevated angle so that the entire stage is visible. A vintage alarm clock stands on the left, a mechanical keyboard occupies the foreground, several thick books are stacked in the middle, a cup of tea and a pad of sticky notes sit beside the books, and an unlit desk lamp stands on the right. These objects retain clearly recognizable everyday shapes, but all are drawn in pixel art. The opening is quiet and clearly organized, leaving enough space between objects for the roads, rivers, and actions that follow.

\emph{First sequence: the alarm clock opens and the star appears.} The alarm clock's two bells suddenly bounce vigorously, sending a ring of square sound waves outward. Its face opens like a mechanical hatch, ejecting a golden-yellow pixel star high into the air along a parabolic arc. A white pixel kitten wearing a blue hat rushes out of a small door at the bottom of the clock, quickly climbs a clock foot, leaps up to catch the star in midair, and lands at the left end of the keyboard. An empty star-shaped socket is visible inside the distant unlit desk lamp, clearly indicating where the kitten must deliver the star. Keep the same kitten, hat, and star throughout all subsequent shots.

\emph{Second sequence: the keyboard becomes a growing railway.} Starlight falls on the keyboard, and keys rise one after another along the route like rows of small buildings. Square tracks assemble piece by piece from the gaps between the keys, extending toward the books. The Enter key rises, and four pixel wheels grow beneath it, transforming it into the kitten's cart. Holding the star, the kitten jumps aboard; the cart immediately accelerates across most of the frame, turning sharply between keys of different heights, descending, and then racing up a ramp. The track continually assembles a short distance ahead of the cart, making the tension clear: the road has only just appeared when the cart races across it. Station flags and windows light up in sequence only after the cart passes, progressively activating the world along its route.

\emph{Third sequence: the spacebar launches the cart and book pages unfold into a slide.} The cart reaches the end of the keyboard and presses down the spacebar. Like a spring-loaded launch pad, the spacebar first visibly sinks and then snaps upward, throwing both cart and kitten toward the books. Show a clear takeoff, airborne trajectory, and landing for this large leap between two objects, rather than instant teleportation. While the kitten is airborne, the top book suddenly opens, and its pages unfold one after another into a long slide winding around the stack. The cart lands on the slide and races downward, rounding bends and jumping short gaps while its wheels lift pages that flutter behind it. The books remain recognizable; their pages have simply become roads within the miniature world.

\emph{Fourth sequence: tea becomes rapids and a sticky note folds into a paper boat.} The unfolded book cover strikes the teacup handle. The cup slowly tilts, then pours out a stream made of light and dark blue pixel blocks. The water first surges past the base of the books and then rushes across the desk toward the lamp, cutting off the original land route. Wave crests consist of constantly rearranging stepped color blocks, while splashes are small squares flying outward. As the kitten reaches the end of the slide, the current lifts a sticky note, which folds continuously in midair into a paper boat. The kitten jumps aboard while holding the star, leaving the cart on the bank. The rapids immediately sweep the boat around the cup's base, down a small waterfall between the books, and upward on a large wave. Give the boat clear plunging, tilting, and airborne movements; it must not merely glide at constant speed on a flat plane.

\emph{Fifth sequence: a pull-cord swing delivers the star to the lamp.} As the paper boat passes beneath the lamp, a pixel pull cord suddenly descends from the shade, assembling segment by segment. The kitten secures the star against its chest, steps onto the bow, leaps high from a wave crest, and grabs the cord. Its momentum draws the cord into a broad pendulum swing: the kitten sweeps through the low point and rises nearly to the top of the frame. At the highest point, it releases the cord, flies along a clear arc into the lampshade, and uses both hands to push the star into the bulb's star-shaped socket. Hold briefly, then let the bulb suddenly light up.

\emph{Ending: the entire desk becomes an illuminated miniature city.} Warm yellow light spreads from the lamp across the desk as successive pixel-color bands with crisp edges, rather than a soft-focus glow. Wherever the light reaches, the world completes its final assembly: keyboard buildings light their windows, railings grow beside the book-page slide, a small dock assembles beside the teacup, and sticky notes fold into birds that fly past the lampshade. The previously turbulent river gradually settles, and the paper boat docks. The kitten sits on the edge of the lampshade, swinging its feet and gazing at the desk city it has illuminated.

Base the animation on action within a continuous space, so viewers always understand where the character starts, which objects it passes, and where it is going. Keep the camera stable, using only slight lateral tracking when necessary; do not rely on rapid cuts or camera shake to create motion. The action climaxes come from launching, sliding, surfing, and swinging; the creativity comes from objects triggering one another, rather than arbitrary scene changes. All new structures must appear through gathering pixel blocks, assembly frame by frame, mechanical unfolding, or paper folding; avoid fading entire structures into view. Keep the pixel grid stable. Do not introduce antialiased edges, motion blur, smooth gradients, realistic water, or blurred glowing edges, and do not add game interfaces, scores, subtitles, or control buttons.

\medskip
\noindent\textbf{The Tiny Dragon Barista.}\par

Build ``The Tiny Dragon Barista'': a 10-second looping cozy fantasy animation, rendered entirely with JavaScript Canvas 2D. No sound, images, spritesheets, or video assets.

\emph{Story.} A realistic-looking morning caf\'{e} counter is overlaid with playful hand-drawn fantasy elements. A coffee cup waits beneath an espresso machine. Sleeping inside the cup is a tiny orange dragon, curled like a cat.

\begin{itemize}
    \item 0.0--1.2 s: steam rises from the espresso machine and tickles the dragon's nose.
    \item 1.2--2.5 s: the dragon sneezes a tiny puff of flame. Comic text: ``pff!''
    \item 2.5--4.0 s: the flame accidentally ignites three floating coffee beans, which become glowing fireflies.
    \item 4.0--5.8 s: the dragon jumps onto the counter and chases them in a circular path around the cup, leaving a hand-drawn orange motion trail.
    \item 5.8--7.2 s: it catches the final bean and breathes one controlled flame onto the coffee foam.
    \item 7.2--8.5 s: the foam transforms into a perfect latte-art dragon face. The tiny dragon proudly places both hands on its hips.
    \item 8.5--10.0 s: too much steam erupts; the dragon dives back into the cup, the foam collapses, and the original sleeping pose returns.
\end{itemize}

\emph{Look.} Combine a semi-realistic caf\'{e} still-life composition with visibly illustrated overlays: inked flames, scribbled steam, glowing beans, exaggerated cartoon expressions.

Palette: espresso brown, cream, muted sage, terracotta orange. Thick dark-brown outlines only on animated fantasy elements, while furniture and countertop use softer realistic shading.

\emph{Details.} Add:
\begin{itemize}
    \item condensation on the ceramic cup,
    \item tiny crumbs moving when the dragon lands,
    \item hand-drawn spark symbols,
    \item a hanging menu gently swaying,
    \item steam morphing into temporary doodles: stars, hearts, spirals.
\end{itemize}

\emph{Loop.} The dragon must finish curled inside the cup in exactly the same pose as frame 0.

\medskip
\noindent\textbf{Watering-Can Rafting.}\par

A real metal watering can sits on a tabletop. Tiny doodle cats arrive carrying paddles, life rings, and signboards, then quickly draw channels, dams, and tiny docks around it. One cat turns the watering can, and a stream of water suddenly pours out. As the water flows, the drawn channels come alive, turning into a roaring doodle river. Little boats appear and launch into the current, carrying cats through sharp turns, splashes, and drops. The flow gets stronger and stronger, creating rapids, waterfalls, and spinning whirlpools made from hand-drawn lines and foam. More doodle fish leap out of the water, boats bounce upward, one cat nearly falls overboard, another surfs on a leaf racing downstream. The entire scene becomes a dramatic flood-powered amusement ride, mixing a real watering can with a vividly animated doodle whitewater world.

\end{document}